%% file: neurips.tex
\documentclass{article}

\input{misc/packages}
\input{misc/commands}

\usepackage[main, final]{neurips_2026}

\usepackage[utf8]{inputenc} 
\usepackage[T1]{fontenc}    
\usepackage[colorlinks=true,linkcolor=black,citecolor=black,urlcolor=black]{hyperref}    
\usepackage{url}            
\usepackage{booktabs}       
\usepackage{amsfonts}       
\usepackage{nicefrac}       
\usepackage{microtype}      
\usepackage{xcolor}         

\title{The Silent Hyperparameter: Quantifying the Impact of Inference Backends on LLM Reproducibility}

\author{%
  David Pape \quad Jonathan Evertz \quad Lea Schönherr \\
  CISPA Helmholtz Center for Information Security, Saarbrücken, Germany\\
  \{\texttt{david.pape, jonathan.evertz, schoenherr\}@cispa.de} \\
}

\begin{document}

\maketitle

\input{content/01_abstract}
\input{content/02_introduction}
\input{content/03_related_work}

\input{content/04_landscape}

\input{content/05_methodology}

\input{content/06_results}
\input{content/07_root_cause}

\input{content/08_paper_survey}
\input{content/09_discussion}
\input{content/10_conclusion}


\newpage

\begin{ack}
This work was supported by the Helmholtz Association's Initiative and Networking Fund on the HAICORE@FZJ partition and by the German Federal Ministry of Research, Technology and Space (BMFTR) under the grants AIgenCY (16KIS2012) and SisWiss (16KIS2330). In addition, this work was funded by zukunft.niedersachsen, the joint science funding program of the Lower Saxony Ministry of Science and Culture and the Volkswagen Foundation, and the Daimler and Benz Foundation under the grant Ladenburger Kolleg, Project KonCheck. \textit{Competing interests: The authors declare no competing interests.} 
\end{ack}

\bibliography{strings, main}
\bibliographystyle{unsrtnat}

\newpage


\appendix

\input{content/12_appendix}


\newpage

\end{document}

%% file: misc/packages.tex
\usepackage{multirow}
\usepackage{siunitx}
\usepackage{bm}
\usepackage{xspace}
\usepackage[inline,shortlabels]{enumitem}
\usepackage{dsfont}
\usepackage{threeparttable}
\usepackage{subcaption}
\usepackage{mwe}
\usepackage{graphicx}
\usepackage[dvipsnames]{xcolor}
\usepackage[most]{tcolorbox}
\usepackage{makecell}
\usepackage{tabularx}
\usepackage{ragged2e}

\usepackage[numbers,sort&compress]{natbib}

\definecolor{taxgray}{gray}{0.55}

%% file: misc/commands.tex
\newcommand{\eg}{e.\,g.,\xspace}


%% file: content/01_abstract.tex
\begin{abstract}
Progress in LLMs is increasingly measured through standardized benchmarks, where state-of-the-art improvements are often separated by fractions of a percentage point. At the same time, the computational cost of evaluating modern LLMs has driven widespread adoption of specialized inference backends, software systems that execute trained models efficiently at inference time. While critical for scalability, system-level optimizations, such as custom CUDA kernels and reduced-precision arithmetic, can alter token probabilities and introduce non-determinism, possibly cascading into divergent generation. 
In this work, we first survey the inference landscape, identifying $200$ distinct engines, and analyze $35,000$ ML publications, finding that the specific inference stack is rarely reported despite this widespread diversity. We then present a systematic empirical study of how inference backends affect LLM benchmark results. Holding model weights, decoding parameters, and hardware constant, we evaluate five widely used inference engines, including vLLM, SGLang, and llama.cpp, across multiple open-weight models spanning 4B-30B parameters and established benchmarks. We show that the choice of backend alone can shift benchmark scores by up to $16.6$ percentage points and induce high rates of output disagreement. By isolating backend optimizations and tracing the execution pipeline, we find this divergence is driven by system-level optimizations such as prefix caching and CUDA graphs, custom kernels, and engine-specific defaults in logit processing. Our findings identify the inference backend as a previously unreported yet consequential hyperparameter in LLM evaluations and advocate standardized reporting of inference stacks to improve the reproducibility and interpretability of benchmark comparisons.
\end{abstract}

%% file: content/02_introduction.tex
\section{Introduction}

The rapid advancement of Large Language Models~(LLMs) has established a new standard in artificial intelligence. However, running evaluations on powerful models introduces significant computational challenges, demanding immense memory and processing power. To address this, a rich ecosystem of specialized inference engines has emerged, including tools such as vLLM~\cite{10.1145/3600006.3613165}, SGLang~\cite{zheng2024sglangefficientexecutionstructured}, and llama.cpp~\cite{llama_cpp}, which are becoming essential for efficient model serving.
These \textit{backends} employ sophisticated optimization techniques, such as paged attention~\cite{10.1145/3600006.3613165}, custom CUDA kernels, and optimized memory management, to reduce latency and increase throughput. Consequently, they are widely adopted not only in production but also by researchers for resource-efficient experimentation.


While essential for performance, these engines are complex systems, and \emph{their internal optimizations can potentially alter model outputs.} Differences in floating-point accumulation, non-deterministic behavior in custom CUDA kernels, or varying implementations of attention mechanisms could lead to subtle numerical differences in token log-probabilities. In the context of autoregressive generation, where the selection of the next token depends on the previous sequence, a single flipped token early in the generation can cascade into a completely divergent output.

This potential source of variance has critical implications. Different inference engines can yield different benchmark scores, even when the underlying model is identical. Consequently, this backend-induced variance can dethrone a model or falsely elevate a weaker one. A model's superior performance might not stem from a better architecture or improved training paradigm, but from the specific numerical properties of the inference engine used during testing. Beyond academic rankings, this instability also poses risks in real-world deployments. A model trained for safety alignment or medical accuracy on a reference implementation (\eg Hugging Face transformers~\cite{wolf-etal-2020-transformers}) may exhibit different, potentially unsafe behaviors when deployed on a high-throughput engine such as vLLM. This creates a dangerous ``deployment gap'' between research validation and production reality, where backend-induced discrepancies can undermine not only performance claims but also safety guarantees, potentially leading to harmful or non-compliant behavior in real-world use.

While prior work has examined sources of variability such as prompt sensitivity \cite{10.1007/978-3-031-88714-7_29}, quantization \cite{kurtic-etal-2025-give}, and decoding strategies \cite{shi-etal-2024-thorough}, the role of the inference backend itself has remained largely unexplored. In this work, we address this gap through a systematic empirical study and a large-scale survey of over 35,000 papers published at top machine learning venues. Our survey reveals that the specific inference stack is rarely reported, despite its widespread use in evaluation and deployment. Complementing this analysis, our controlled experiments demonstrate that the choice of inference backend alone can induce substantial variation in benchmark outcomes, shifting reported performance by up to $16.6$ percentage points, even when model weights, prompts, and decoding parameters are held constant.
 
In summary, our contributions are as follows:
\begin{itemize}[left=0pt, topsep=0pt]
    \item \textbf{Landscape Survey.} We survey the landscape of modern inference engines and categorize them.
    \item \textbf{Controlled Evaluation.} We conduct a unified evaluation of open-weight models across a diverse set of popular backends (including vLLM, SGLang, and llama.cpp). We quantify their differences on standard benchmarks, demonstrating that the choice of backend is a significant hyperparameter.
    \item \textbf{Root Cause Analysis.} By isolating specific optimizations, we trace backend-induced variance to two primary sources: correctable systematic defaults and optimization-induced numerical drift.
    \item \textbf{Reproducibility in ML Research.} We analyze over $35,000$ recent publications from top ML conferences to quantify how frequently the inference stack is reported. 
\end{itemize}
By quantifying this variability, we aim to establish new reporting standards that improve scientific reproducibility in the era of optimized inference. To support reproducibility and future research, we release all code and experimental artifacts at \url{https://github.com/Dormant-Neurons/inference_backends}.

%% file: content/03_related_work.tex
\section{Related Work}
Our work connects to a broad literature on LLM inference and reproducibility.

\textbf{Reproducibility for LLM evaluations.} Recent studies highlight varying LLM results due to floating-point non-associativity~\cite{yuan2025understanding}, ambiguous semantic benchmarks~\cite{Biderman2024LessonsFT}, model versioning~\cite{evertz-26-shadows}, request-routing services~\cite{higton2026routing}, and serving-stack effects on tool-use evaluation~\cite{tang2026serving}. We extend this literature by isolating the inference engine itself as a key, previously undocumented source of evaluation variance.



\textbf{Assessing inference performance and determinism.} Prior evaluations of inference engines focus primarily on hardware optimizations, speed, and energy efficiency~\cite{10820566, li2024large, park2025surveyinferenceengineslarge}. Other work studies system-design choices such as caching and decoding~\cite{2025miaotowardsefficient}, extends earlier findings on CNNs and RNNs~\cite{xu2026hardwareaccelerationneuralnetworks}, or examines quantization-induced differences~\cite{wang2026hiddenreliabilityriskslarge}, including cross-kernel scaling and rounding effects~\cite{chen2026integeralibi}. Recent security studies further exploit platform- and kernel-dependent numerical differences to activate backdoors in LLMs and VLMs~\cite{loose2026floatdoorplatformtriggeredbackdoorsllms, cheng2026vlmsystrojan}. Despite optimizations for determinism~\cite{btad164} and guidelines recommending fixed seeds and low temperatures~\cite{Blackwell2024TowardsRL}, LLMs retain inherent randomness. We build on these performance-centric studies to quantify how backend design choices fundamentally alter generation trajectories.

%% file: content/04_landscape.tex
\section{Landscape of Modern Inference Engines}\label{sec:landscape_survey}
LLM inference is resource-intensive and latency-sensitive, requiring careful management of memory, parallelism, and hardware utilization.
Inference engines encapsulate these optimizations behind standardized execution interfaces. We survey this ecosystem to quantify its diversity.

\begin{figure}[!t]
    \centering
    \begin{subfigure}[b]{0.48\textwidth}
        \centering
        \includegraphics[width=\textwidth, trim=0 0.25cm 0 0.25cm, clip]{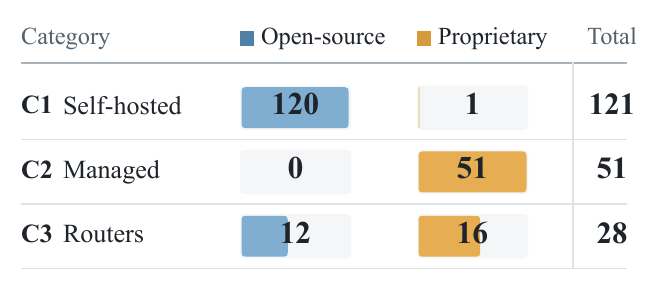}
        \caption{Engine Categories}
        \label{fig:landscape_survey}
    \end{subfigure}
    \hfill
    \begin{subfigure}[b]{0.48\textwidth}
        \centering
        \includegraphics[width=\textwidth, trim=0 0.25cm 0 0.25cm, clip]{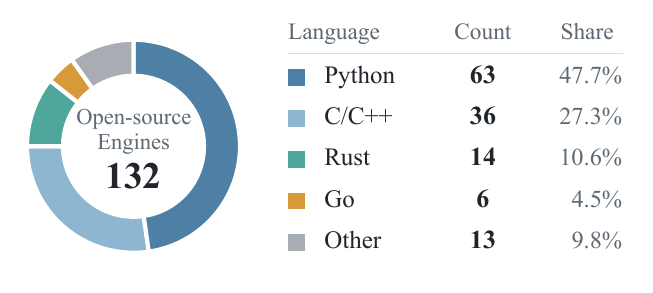}
        \caption{Programming Languages}
        \label{fig:programming_distribution}
    \end{subfigure}
    
    \caption{\textbf{Landscape of Inference Engines.} \textbf{(a)} Counts of the $200$ surveyed inference engines by category (C1-C3) and source-code availability (blue: open-source; orange: proprietary). \textbf{(b)} The distribution of primary programming languages used across the open-source engines.} 
    \label{fig:combined_landscape}
\end{figure}

\subsection{Survey Methodology and Scope}
We define an \textit{inference engine} as standalone software that can load a transformer-based model and generate completions. We surveyed the ecosystem as of January 2026.

\textbf{Search Strategy.} We identified candidates through GitHub searches (\eg ``LLM inference'' and ``inference-engine''), general web search, and community discussions on platforms including Reddit and Hugging Face.

\textbf{Inclusion and Exclusion Criteria.} We include software that (1) supports open-weight or local models, (2) has a verifiable open-source repository or API, and (3) shows evidence of adoption or availability (at least 100 GitHub stars for open-source projects or an active API for proprietary services). These criteria focus the survey on systems relevant to open-weight evaluation that can be independently identified and accessed. We exclude foundational libraries (\eg PyTorch, JAX) and training/application-layer tools (\eg LangChain), as neither is a dedicated inference engine. We further exclude engines that serve only VLMs or diffusion models, as they fall outside our text-generation scope.


\subsection{Taxonomy of Inference Systems}\label{subsec:landscape_taxonomy}
We classify these engines based on the level of control a user holds over the hardware and software environment:
\par\smallskip

\begingroup
\footnotesize
\setlength{\fboxsep}{6pt}
\setlength{\fboxrule}{0.45pt}

\newcommand{\onecard}[4]{%
\fcolorbox{black!25}{black!2}{%
\begin{minipage}[t][2.5cm][t]{0.29\linewidth}
\RaggedRight
{\color{taxgray}\bfseries #1}\par\vspace{3pt}
\textbf{#2}\par\vspace{1pt}
#3\par\vfill
\textit{Examples:}\par #4
\end{minipage}}}

\noindent
\onecard{Category 1}
{Self-hosted inference libraries}
{Users run the model and manage inference software and compute.}
{vLLM, llama.cpp, SGLang.}%
\hfill
\onecard{Category 2}
{Managed inference platforms}
{Providers run models and expose inference through APIs.}
{Fireworks AI, Together AI.}%
\hfill
\onecard{Category 3}
{Aggregators and routers}
{A interface routes requests to providers or self-hosted endpoints.}
{OpenRouter, LiteLLM.}

\par
\endgroup
\subsection{Landscape Analysis}
Our search identified $\mathbf{200}$ inference engines (Figure~\ref{fig:landscape_survey}): $121$ self-hosted libraries (Category~1; $60.5$\,\%), $51$ managed platforms (Category~2; $25.5$\,\%), and $28$ aggregators/routers (Category~3; $14.0$\,\%). This variety of available engines shows software choice is a significant variable in experiments. Of these, $132$ are open-source, concentrated in Category~1 ($120$/$121$); all Category~2 systems are proprietary, while Category~3 includes both ($12$ open-source, $16$ proprietary). Source availability affects how much of each system can be inspected and managed by the user.\\However, we find that $44$ engines ($22$\,\% of the total) are inactive (no main branch commits in six months); $43$ of these are self-hosted projects, with only one belonging to Category~3. This highlights that maintaining complex, low-level LLM optimizations quickly outpaces individual developer capacity. Finally, we analyzed the primary programming languages for the open-source subset of our survey (Figure~\ref{fig:programming_distribution}). While Python remains the dominant engine language, many backends are implemented in C/C++ or Rust, reflecting the necessity of low-level languages for efficient model serving.

%% file: content/05_methodology.tex
\section{Methodology}\label{sec:methodology}
To quantify the impact of inference backends on model reproducibility, we isolate inference engine effects using a controlled experimental framework. 

\subsection{Experimental Scope}

\textbf{Inference Backends.} We selected five Category 1 (self-hosted) inference engines based on adoption and experimental compatibility: vLLM~\cite{10.1145/3600006.3613165}, SGLang~\cite{zheng2024sglangefficientexecutionstructured}, llama.cpp~\cite{llama_cpp}, LMDeploy~\cite{2023lmdeploy, zhang2025efficient}, and Ollama~\cite{ollama}. We use Hugging Face transformers~\cite{wolf-etal-2020-transformers} as the reference implementation. Self-hosting lets us hold the surrounding inference environment fixed and isolate the backend; managed platforms and routers introduce additional hidden or provider-controlled components. Full backend-selection details are provided in Appendix~\ref{app:backend_versions}.

\textbf{Models.} We selected five open-weight text-generation models spanning 4B-30B parameters and multiple architectures. \textit{Standard models: } Llama 3.1 8B~\cite{grattafiori2024llama3herdmodels}, Qwen3 4B, Qwen3 30B~\cite{qwen3}. \textit{Reasoning models:} DeepSeek R1 Distill Qwen 7B~\cite{Guo_2025} and Qwen3 Thinking 30B~\cite{qwen3}. 

\textbf{Benchmarks.} To study performance differences across inference backends rather than absolute model performance, we use four established benchmarks covering distinct tasks: GSM8K (Math)~\cite{Cobbe2021TrainingVT}, GPQA Diamond (Science)~\cite{rein2024gpqa}, SimpleQA Verified (Factuality)~\cite{haas2025simpleqaverifiedreliablefactuality}, and LiveCodeBench v6 (Code)~\cite{jain2024livecodebench}. 

Licensing information for the models and datasets used in this study is provided in Appendix~\ref{app:licenses}.

\subsection{Standardization}\label{subsec:standardization}
To attribute performance differences strictly to the backend implementation, we enforce the following constraints:

\begin{itemize}[left=0pt, label={},topsep=0pt,leftmargin=0pt]
    \item \textbf{Decoding Strategy.} We enforce greedy decoding ($\text{temperature}=0$) for all generations.
    \item \textbf{Model Precision.} All models are loaded in FP16 (GGUF-FP16 for llama.cpp/Ollama).\footnote{Not all engines support FP32, and FP16 serves as a baseline across all evaluated engines.} 
    \item \textbf{Prompting.} We apply the tokenizer's Jinja2 chat template externally to avoid backend-specific templating.
    \item \textbf{Batch Size.} Evaluations use a batch size of one to prevent batching-induced instability~\cite{batch_nondeterminsm,yuan2025understanding}.
    \item \textbf{Generation Parameters.} We set the maximum number of output tokens to $2,048$ (Standard) and $32,768$ (Reasoning), with context windows of $4,096$ and $34,816$ respectively. 
    \item \textbf{Seeds.} To account for other sources of non-determinism, we report averages across twelve unique seeds for every configuration.
\end{itemize}

%% file: content/06_results.tex
\section{Evaluation and Results}\label{sec:eval}

All evaluations use a unified Docker container (Ubuntu 22.04, Python 3.12), a single NVIDIA H100 GPU, and fixed backend versions (Appendix~\ref{app:backend_versions}, Table~\ref{tab:backend_versions}). We study behavioral reproducibility by combining aggregate scores with per-example, token-level, and probability-level divergence. Complementary throughput and latency measurements are reported in Appendix~\ref{app:efficiency_metrics}.

\subsection{Outcome Consistency}\label{subsec:eval_outcome}
To measure macro-level agreement, we evaluate Benchmark Accuracy, Disagreement Rate, and Length Error. 
\begin{itemize}[left=0pt, topsep=0pt, label={},leftmargin=0pt]
\item \textbf{Benchmark Accuracy (Table~\ref{tab:comparison_experiment}).} Our results indicate that the inference engine is a significant source of variance, with accuracy discrepancies often exceeding several percentage points, enough to alter leaderboard rankings. Llama 3.1 8B on Ollama exhibits a sharp performance drop on GSM8K, falling ten percentage points below the reference. The impact is most pronounced in reasoning models. DeepSeek R1 Distill Qwen 7B displays a $16.60$ percentage-point spread between the best- and worst-performing backends on GSM8K.\\We additionally examine whether backend-induced score shifts change model ordering. Across our model set, $19$ of $20$ backend-benchmark combinations preserve the transformers ordering, which is unsurprising when models are separated by large performance gaps. However, for more closely matched models, reversals can occur: on GPQA, LMDeploy reverses Qwen3 4B and DeepSeek R1 Distill Qwen 7B relative to transformers ($35.35$ vs $35.86$, compared with $35.86$ vs $33.33$). This suggests that backend choice is particularly consequential when comparing models with similar benchmark performance.\\ Ultimately, no backend matches transformers across all benchmarks. Nor do we find evidence for a universally superior backend. Across the $20$ model-benchmark combinations, four backends achieve the best mean rank for at least one model; Appendix~\ref{app:best_backend} provides the full analysis. We therefore treat backend selection as an experimental variable rather than an optimization for accuracy.

\item \textbf{Disagreement Rate (Figure~\ref{fig:disagreement_rate} in Appendix~\ref{app:diasgreement}).} We define this as the frequency with which a backend's prediction $y$ differs from the reference $y_{ref}$ for the same input, regardless of ground-truth correctness: $D = \frac{1}{N}\sum\mathds{1}(y \neq y_{ref})$. While in some settings the disagreement is small, in many cases we observe that disagreement rates consistently exceed the absolute differences in accuracy, indicating that the backend alters the model's decision boundary. For example, Ollama disagrees with transformers on $27.37$\,\% of DeepSeek R1 Distill Qwen 7B's GSM8K predictions, more than $360$ of $1,319$ questions. Disagreement is also pronounced for reasoning models on GPQA and LiveCodeBench. 

\item \textbf{Length Error (Figure~\ref{fig:length_error} in Appendix~\ref{app:length_error}).} We detect systematic biases in verbosity via the signed difference (bias) and absolute difference (magnitude) in output token counts.  Standard models usually stay near a stable ``ideal zone'' of $\leq$$25$ absolute tokens, a threshold indicating the text length varies by no more than 1-2 sentences, preserving structural consistency. However, the lengths for GPQA and LiveCodeBench differ substantially for all tested models. Reasoning models show large length differences on all datasets. For instance, the Ollama backend produces DeepSeek R1 Distill Qwen 7B outputs that are, on average, over $9,000$ tokens shorter than the reference on GPQA, fundamentally altering the chain-of-thought process. 

\end{itemize}

\begin{table*}[!t] 
\centering
\caption{\textbf{Backend Performance Variance across Benchmarks.} Performance of five inference backends and the transformers reference across models and benchmarks. Metrics are accuracy (\%) for GPQA and GSM8K, F1 for SimpleQA, and pass@1 for LiveCodeBench. The final column reports the maximum$-$minimum score difference. Highest scores are \textbf{bold}; lowest scores are \underline{underlined}.}
\label{tab:comparison_experiment}
\resizebox{\textwidth}{!}{%
\begin{tabular}{llc|ccccc|c}
\toprule
\multirow{2}{*}{\textbf{Model}} & \multirow{2}{*}{\textbf{Benchmark}} & 
\textbf{transformers} & \multirow{2}{*}{\textbf{llama.cpp}} & \multirow{2}{*}{\textbf{LMDeploy}} & \multirow{2}{*}{\textbf{Ollama}} & \multirow{2}{*}{\textbf{SGLang}} &  \multirow{2}{*}{\textbf{vLLM}} & \multirow{2}{*}{\textbf{Max - Min}} \\
 &  & 
\multicolumn{1}{c|}{\textbf{(reference)}} & & & & &  & \\
\midrule

\multirow{3}{*}{Qwen3 4B}
& GPQA & 35.86 $\pm$ 00.00 & \textbf{38.89} $\pm$ 00.00 & 35.35 $\pm$ 00.00 & 37.88 $\pm$ 00.00 & 34.85 $\pm$ 00.00 &  \underline{34.81} $\pm$ 00.55 & 04.08 \\
& GSM8K &  \textbf{90.83} $\pm$ 00.00 & \underline{90.45} $\pm$ 00.00 & \underline{90.45} $\pm$ 00.00 & 90.75 $\pm$ 00.00 & \underline{90.45} $\pm$ 00.00 & 90.47 $\pm$ 00.15 & 00.38 \\
& SimpleQA & \textbf{05.91} $\pm$ 00.09 & 05.79 $\pm$ 00.07 & 05.89 $\pm$ 00.10 & 05.81 $\pm$ 00.09 & \underline{05.52} $\pm$ 00.06 &  05.83 $\pm$ 00.09 & 00.40 \\
& LiveCodeBench &  29.71 $\pm$ 00.00 & 30.29 $\pm$ 00.00 & 30.86 $\pm$ 00.00 & \underline{29.14} $\pm$ 00.00 & 30.29 $\pm$ 00.00 & \textbf{32.33} $\pm$ 00.38 & 03.19 \\
\midrule

\multirow{3}{*}{Llama 3.1 8B}
& GPQA & 24.75 $\pm$ 00.00 &  23.74 $\pm$ 00.00 & \textbf{25.76} $\pm$ 00.00 & \underline{22.22} $\pm$ 00.00 & 23.23 $\pm$ 00.00 & \textbf{25.76} $\pm$ 00.00 & 03.54 \\
& GSM8K & 84.23 $\pm$ 00.00 & 84.15 $\pm$ 00.00 & \textbf{84.53} $\pm$ 00.00 & \underline{74.30} $\pm$ 00.00 & 84.23 $\pm$ 00.00 &  83.85 $\pm$ 00.00 & 10.24 \\
& SimpleQA & \underline{01.81} $\pm$ 00.12 & 01.97 $\pm$ 00.15 & 01.89 $\pm$ 00.23 & \textbf{02.64} $\pm$ 00.12 & 01.84 $\pm$ 00.13 &  02.06 $\pm$ 00.11 & 00.83 \\
& LiveCodeBench & 17.71 $\pm$ 00.00 & 17.71 $\pm$ 00.00 & \textbf{18.29} $\pm$ 00.00 & \underline{13.14} $\pm$ 00.00 & \textbf{18.29} $\pm$ 00.00 &  \textbf{18.29} $\pm$ 00.00 & 05.14 \\
\midrule 

\multirow{3}{*}{Qwen3 30B}
& GPQA & \textbf{45.45} $\pm$ 00.00 & 42.42 $\pm$ 00.00 & 40.40 $\pm$ 00.53 & \textbf{45.45} $\pm$ 00.00 & 41.41 $\pm$ 00.00 & \underline{40.15} $\pm$ 01.44 & 05.30 \\
& GSM8K & 91.74 $\pm$ 00.00 & 91.74 $\pm$ 00.00 & 91.79 $\pm$ 00.10 & \textbf{91.81} $\pm$ 00.00 & 91.74 $\pm$ 00.00 & \underline{91.72} $\pm$ 00.10 & 00.09 \\
& SimpleQA & \underline{22.63} $\pm$ 00.10 & 22.98 $\pm$ 00.13 & 22.82 $\pm$ 00.11 & 22.91 $\pm$ 00.11 & \textbf{23.00} $\pm$ 00.08 & 22.66 $\pm$ 00.18 & 00.38 \\
& LiveCodeBench & 38.29 $\pm$ 00.00 & 37.14 $\pm$ 00.00 & \underline{36.57} $\pm$ 00.84 & 38.86 $\pm$ 00.00 & 37.71 $\pm$ 00.00 &  \textbf{39.14} $\pm$ 01.55 & 02.57 \\
\midrule
\midrule

\multirow{3}{*}{\shortstack{DeepSeek R1 \\ Distill Qwen 7B}}
& GPQA & 33.33 $\pm$ 00.00 & 31.82 $\pm$ 00.00 & \textbf{35.86} $\pm$ 00.00 & \underline{27.27} $\pm$ 00.00 & 31.31 $\pm$ 00.00 & 33.84 $\pm$ 00.00 & 08.59 \\
& GSM8K & \textbf{78.47} $\pm$ 00.00 & 78.24 $\pm$ 00.00 & 73.54 $\pm$ 00.00 & \underline{61.87} $\pm$ 00.00 & 78.24 $\pm$ 00.00 &  78.32 $\pm$ 00.00 & 16.60 \\
& SimpleQA & \textbf{04.88} $\pm$ 00.14 & \underline{03.88} $\pm$ 00.24 & 04.36 $\pm$ 00.08 & 04.86 $\pm$ 00.19 & 04.36 $\pm$ 00.18 &  04.74 $\pm$ 00.19 & 01.00 \\
& LiveCodeBench & 20.57 $\pm$ 00.00 & 22.86 $\pm$ 00.00 & 22.29 $\pm$ 00.00 & \underline{19.43} $\pm$ 00.00 & 22.29 $\pm$ 00.00 & \textbf{25.14} $\pm$ 00.00 & 05.71 \\
\midrule

\multirow{3}{*}{\shortstack{Qwen3 \\Thinking 30B}}
& GPQA & 71.72 $\pm$ 00.00 & 70.71 $\pm$ 00.00 & 69.70 $\pm$ 00.00 & 72.22 $\pm$ 00.00 & \textbf{72.73} $\pm$ 00.00 &  \underline{69.28} $\pm$ 02.56 & 03.45 \\
& GSM8K & \textbf{94.39} $\pm$ 00.00 & 94.24 $\pm$ 00.00 & \underline{94.01} $\pm$ 00.11 & \underline{94.01} $\pm$ 00.00 & 94.31 $\pm$ 00.00 &  94.16 $\pm$ 00.15 & 00.38 \\
& SimpleQA & \textbf{28.71} $\pm$ 00.20 & 28.62 $\pm$ 00.28 & 27.65 $\pm$ 00.32 & \underline{27.33} $\pm$ 00.34 & 27.99 $\pm$ 00.24 &  28.43 $\pm$ 00.36 & 01.37 \\
& LiveCodeBench & \textbf{61.71} $\pm$ 00.00 & 57.71 $\pm$ 00.00 & 60.52 $\pm$ 01.47 & \underline{56.00} $\pm$ 00.00 & 61.14 $\pm$ 00.00 &  60.33 $\pm$ 01.15 & 05.71 \\

\bottomrule
\end{tabular}
} 
\end{table*}

\subsection{Token-Level Divergence}
To pinpoint \textit{when} the generation differences occur, we define the \emph{\textbf{Divergence Index}} as the position $k$ of the first mismatched token between $y$ and $y_{ref}$. We report both the averaged raw index and a Normalized Divergence Score, calculated as $S = \frac{k}{max(|y|, |y_{ref}|)} \in [0,1]$, where $1.0$ indicates a perfect match, while values approaching $0.0$ denote immediate divergence. Analyzing this index (Figures~\ref{fig:divergence_gpqa}--\ref{fig:divergence_livecodebench} in Appendix~\ref{app:divergence}), we observe that reasoning models consistently diverge from the reference generation, much earlier than standard models. This is amplified on difficult benchmarks like GPQA, where, for example, Llama 3.1 8B served via Ollama diverges as early as the 12th output token.


\subsection{Numerical Precision}
To evaluate floating-point stability on the matching prefix (tokens generated prior to divergence), we compute two probability metrics. \emph{\textbf{Logprob Root Mean Squared Error~(RMSE)}} of the top-1 token measures absolute floating-point drift, while \emph{\textbf{Top-5 Token Jaccard Similarity}} assesses preservation of the distribution's  overall shape, even when the top-selected token remains identical. Even small differences are consequential; an RMSE of $0.1$ corresponds to a roughly $10$\,\% relative change in the raw probability assignment ($e^{0.1}\approx1.1$), which is sufficient to flip the greedy selection. While most backends maintain similar precision with the reference (RMSE$<$$0.01$), we observe that reasoning models systematically exhibit higher drift and specific configurations display distinct error spikes. Despite these numerical fluctuations, the Top-5 Token Jaccard similarity remains high across most configurations, suggesting that the general \textit{shape} of the probability distribution remains intact, breaking down only in the most extreme failure cases.

\subsection{Robustness and Real-World Implications}
To ensure our findings generalize beyond our strictly standardized setup, we conducted targeted ablation studies (full details in Appendix~\ref{app:ablations}).
\begin{itemize}[left=0pt, itemsep=0pt, topsep=0pt, label={},leftmargin=0pt]
    \item \textbf{Safety implications (Appendix~\ref{app:safety}).} Using JailbreakBench~\cite{chao2024jailbreakbench}, we found that DeepSeek R1 Distill Qwen 7B's vulnerability to adversarial prompts fluctuated by 8.9 percentage points based solely on the inference engine. 

    \item \textbf{Batching (Appendix~\ref{app:batching}).} Evaluating batched generation (batch size = 4) revealed that performance differences persist, and slight numerical shifts within vLLM and SGLang verify that batching actively influences generation.

    \item \textbf{Hardware Variation (Appendix~\ref{app:hardware}).} Evaluating on NVIDIA L40 GPUs yielded consistent, slightly increased backend-induced variance (up to 17.2 percentage points between backends), confirming divergence is rooted in software implementations rather than specific GPU architectures.

    \item \textbf{Stochastic Sampling (Appendix~\ref{app:sampling}).} Relaxing our greedy decoding constraint to use temperature sampling ($T = 0.7$) preserves the backend-induced variance, proving this variance is not an artifact of greedy decoding.
\end{itemize}

%% file: content/07_root_cause.tex
\section{Root Cause Analysis of Backend Variance}\label{sec:root_cause}
To explain the score shifts in Section~\ref{sec:eval}, we ablate specific optimizations and trace execution across Llama 3.1 8B, DeepSeek R1 Distill Qwen 7B, and Qwen3 4B on GSM8K and GPQA. We categorize the observed causes into two groups: (1) systematic, correctable engine defaults that alter the prompt or decoding behavior and can explain some of the largest performance divergences observed in Table~\ref{tab:comparison_experiment}, and (2) optimization-induced numerical drift that inherently alters the mathematical execution of the model (detailed ablation results are provided in Appendix~\ref{app:root_cause}).

\subsection{Systematic Engine Defaults}\label{subsec:engine_defaults}
The largest effects in our ablation study arise from engine-specific defaults when these are applicable. In particular, several of the extreme divergences in Table~\ref{tab:comparison_experiment}, such as DeepSeek R1 Distill Qwen 7B scoring only $61.87$\,\% on Ollama compared to $78.47$\,\% on the transformers reference, can be traced to hidden changes to the prompt or decoding procedure rather than numerical optimizations alone. 
\begin{itemize}[left=0pt, itemsep=0pt, topsep=0pt, label={},leftmargin=0pt]
\item \textbf{Hidden Prompt Mutation:} Even with an externally applied Jinja2 template and Ollama's \texttt{raw=True} parameter, Ollama prepends a beginning-of-sequence (BOS) token; LMDeploy also inserts a BOS token for DeepSeek R1 Distill Qwen 7B. Removing these tokens raises GSM8K accuracy by $8.34$ points for Llama 3.1 8B with Ollama, $7.35$ for DeepSeek R1 Distill Qwen 7B with Ollama, and $4.70$ for DeepSeek R1 Distill Qwen 7B with LMDeploy. On GPQA, removing Ollama's BOS token raises Llama 3.1 8B accuracy by $4.55$ points. Qwen3 4B is unaffected because its template does not use a BOS token.
\item \textbf{Default Penalty Parameters:} Ollama applies a default repetition penalty of $1.1$, severely degrading chain-of-thought generation. Setting it to $1.0$ raises GSM8K accuracy by $11.67$ points for DeepSeek R1 Distill Qwen 7B and $1.67$ for Llama 3.1 8B, substantially narrowing the largest gaps. On GPQA, it changes accuracy by +$2.02$ points for Llama 3.1 8B and +$1.52$ for DeepSeek R1 Distill Qwen 7B, with no aggregate effects for Qwen3 4B.
\end{itemize}

The defaults can therefore explain large backend-induced differences, but their effects depend on the model, backend, and benchmark. Unlike the numerical effects below, they are largely correctable once prompt tokens and decoding parameters are aligned.

\subsection{Optimization-Induced Numerical Variance}
After aligning prompts and decoding defaults, backend-specific optimizations still introduce numerical differences that can alter generation; their magnitude and direction vary across settings.

\begin{itemize}[left=0pt, itemsep=0pt, topsep=0pt, label={},leftmargin=0pt]
\item \textbf{Prefix Caching and CUDA Graphs:} vLLM, SGLang, llama.cpp, and Ollama enable prefix caching by default. Processing a prompt in fragmented chunks changes floating-point reduction order and can therefore alter subsequent computations. Disabling prefix caching ranges  from negligible effects to several percentage points. On GPQA, it changes DeepSeek R1 Distill Qwen 7B's accuracy by +$3.54$ points with Ollama and $-$$0.51$ with llama.cpp.\footnote{GPQA contains $198$ examples, so one net change in correctness corresponds to approximately $0.51$ percentage points.} CUDA graphs show a similarly configuration-dependent effect: disabling them leaves aggregate accuracy unchanged in several settings, while shifting GPQA accuracy by up to $1.52$ points.
\item \textbf{Kernel-Level Tie-Breaking and Accumulation Precision:} When tokens have identical FP16 logits, PyTorch deterministically selects the lowest token ID, whereas LMDeploy performs greedy decoding through a parallel Top-K kernel ($K=1$) creating a hardware race condition that picks arbitrarily. Patching it to match PyTorch changes DeepSeek R1 Distill Qwen 7B's GPQA accuracy by $-$$3.54$ points, although other settings change little or not at all. llama.cpp and Ollama also accumulate intermediate matrix multiplications in FP32, introducing additional rounding differences relative to pure FP16 execution.
\item \textbf{Layer-wise Error Propagation:} To verify if a specific architectural layer was responsible for these divergences, we implemented a custom tracking pipeline to measure similarity at every layer boundary during the forward pass. We found no single failing layer. Instead, slight numerical drifts caused by custom kernels accumulate across layers, eventually cascading into different Top-1 token predictions.
\end{itemize}

Overall, unlike the engine defaults in Section~\ref{subsec:engine_defaults}, we find no individual numerical optimization that consistently increases or decreases performance across settings.

%% file: content/08_paper_survey.tex
\section{Inference Reproducibility in ML Research}\label{sec:paper_survey}
Given the substantial impact of backend choice on benchmark performance, we conducted a systematic survey of recent ML publications to contextualize this variance and measure the prevalence of inference stack reporting.

\subsection{Methodology}
\textbf{Paper Corpus.} We collected accepted papers published from 2023-2025 at top-tier ML and NLP conferences (NeurIPS, ICML, ICLR, ACL, and EMNLP) via the OpenReview and ACL Anthology APIs. This yielded an initial corpus of more than $35,000$ papers.

\textbf{Reproducibility Tiers.} To categorize reproducibility artifacts available for each paper, we classified papers into four Reproducibility Tiers: \textbf{Tier 0}, neither an inference backend nor a code artifact is mentioned; \textbf{Tier 1}, the backend is documented textually, but no code artifact is provided; \textbf{Tier 2}, an inspectable code repository is available, but specific environment dependencies are absent; and \textbf{Tier 3}, an inspectable code repository is provided alongside an explicit dependency or environment specification, such as \texttt{requirements.txt}, \texttt{pyproject.toml}, or a \texttt{Dockerfile}.\\These tiers distinguish between verifiable and machine-readable environment definitions (Tier 3) and ambiguous textual descriptions that fail to capture implementation details (Tier 1). While dependency files do not guarantee strict reproducibility (\eg unpinned versions), their presence serves as a proxy for reproducibility intent, distinguishing raw code dumps from deliberate standardization efforts.

\textbf{Filtering and Extraction Pipeline.}
We extracted PDF text using \texttt{pymupdf} and applied a keyword pre-filter to identify potentially relevant papers. An LLM-based relevance classifier then identified $9,018$ papers in which the authors perform local inference with open-weight text-generation models. For these papers, we ran two parallel extraction processes using the LLM judge: a \emph{Text Analysis} scan to determine if specific inference engines were explicitly named, and a \emph{Code Analysis} scan to extract repository URLs. Appendix~\ref{app:paper_survey_pipeline} details the pipeline; \ref{app:paper_survey_keywords}-~\ref{app:paper_survey_judge_prompts} provide keywords and judge prompts.

\textbf{Repository Inspection.} For direct GitHub repository URLs, we used the GitHub API to inspect repository contents systematically for dependency and environment specifications (Appendix~\ref{app:dep_files} provides the complete set of file patterns). Code and project URLs that did not directly resolve to GitHub repositories were still counted as available artifacts, but their environment specifications were not assessed systematically. Consequently, these artifacts cannot be reliably distinguished between Tier~2 and Tier~3 based on our automated analysis. 


\begin{figure}[!t]
    \centering
    \includegraphics[width=\textwidth, trim=0 0.75cm 0 0.75cm, clip]{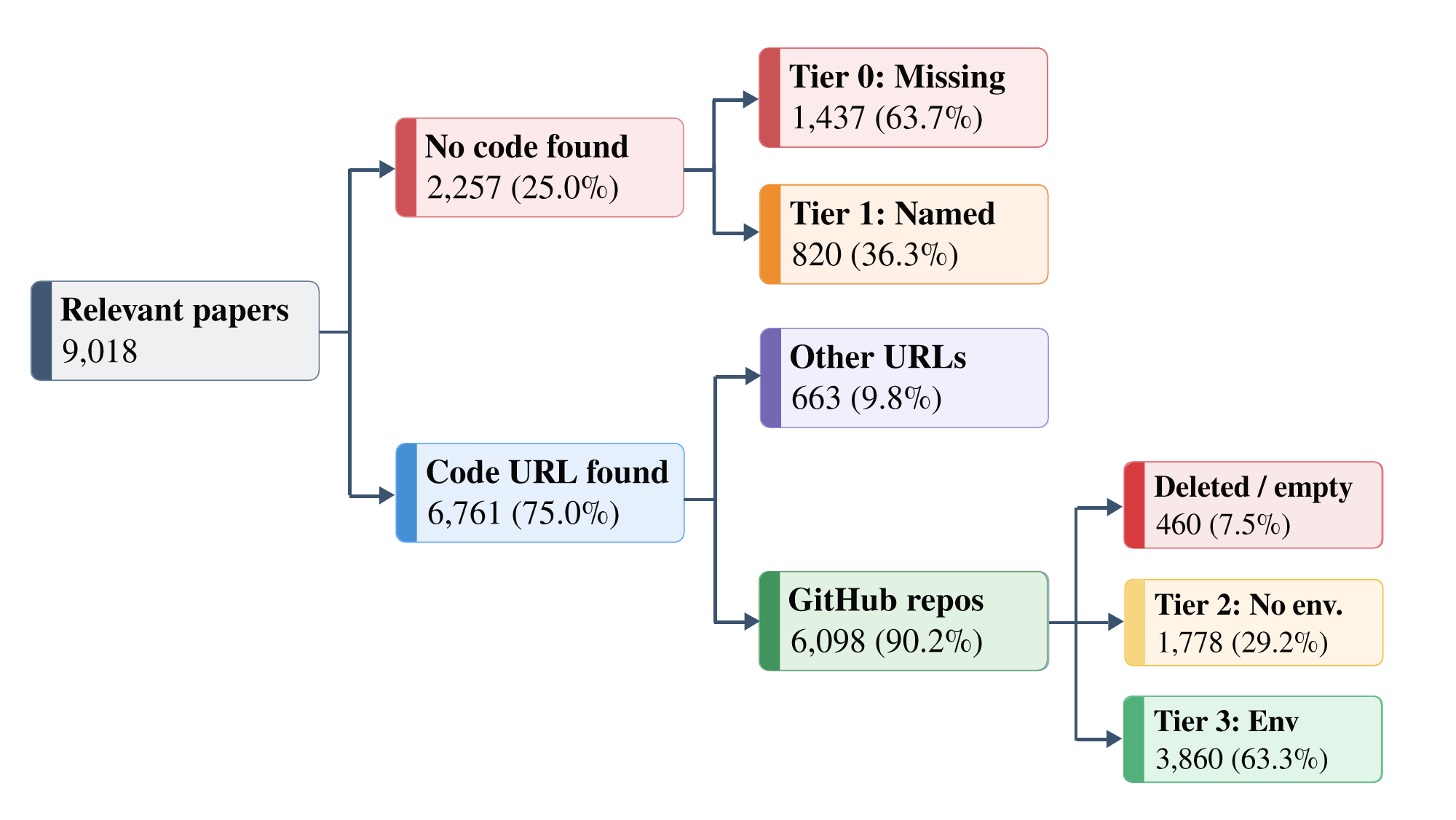} 
    \caption{\textbf{Reproducibility artifacts among $9,018$ papers using local LLM inference.} Percentages are relative to the parent node. Tiers 2 and 3 are assigned only to inspectable GitHub repositories.}
    \label{fig:paper_survey}
\end{figure}

\subsection{Results}

\textbf{Setup.} We utilized Qwen3-235B-A22B-Instruct-2507-AWQ~\cite{Qwen3-235B} as the LLM judge, served with vLLM (version 0.13.0) using greedy decoding.

We present the distribution of reproducibility tiers across the $9,018$ relevant papers in Figure~\ref{fig:paper_survey}.

\textbf{Code Artifact Availability.} Of the $9,018$ relevant papers, $6,761$ ($75.0$\,\%) contained an author-provided code or project URL. Of these, $6,098$ ($90.2$\,\%) were direct GitHub repositories; the remaining $663$ URLs comprised $247$ ($3.7$\,\%) GitHub Pages, $74$ ($1.1$\,\%) Hugging Face URLs, and $342$ ($5.0$\,\%) URLs hosted elsewhere. These were counted as available artifacts, but their environment specifications were not assessed systematically. Among the $6,098$ GitHub repositories, $460$ were deleted, empty, or contained only documentation (\eg a README or license) with no source code.

\textbf{Backend Reporting Frequency.} For papers without code artifacts, we analyzed how frequently the inference stacks were disclosed. Among the $2,257$ papers offering no code, $820$ ($36$\,\%) explicitly named the backend. This subset is dominated by transformers ($322$; $39$\,\%) and vLLM ($150$; $18$\,\%), followed by custom PyTorch implementations ($98$; $12$\,\%). We extended this analysis to the $460$ papers with empty or deleted repositories, and found a similar trend: only $180$ ($39.1$\,\%) reported the engine textually. In this group, reliance on transformers was even higher ($90$; $50$\,\%), with vLLM and custom PyTorch both at $14$\,\% ($25$). 

\textbf{Environment Reproducibility.} We found that only $3,860$ of the accessible GitHub repositories ($\approx$ $63$\,\%) contained explicit dependency or environment specifications. This leaves over a third ($1,778$) of released repositories without a defined execution environment. For these papers, it becomes impossible to reconstruct the specific inference stack used during evaluation.

\textbf{Manual Verification.} To ensure the reliability of our automated pipeline, we conducted a manual verification on a random subset of papers at each filtering and classification stage (details in Appendix~\ref{app:verification}).

Ultimately, these findings confirm that while code sharing is becoming standard, the inference backend remains a largely undocumented source of experimental variance.

%% file: content/09_discussion.tex
\section{Discussion}

The findings above demonstrate that inference backends are not a trivial implementation detail, but an active, influential element of the LLM evaluation process.

\textbf{Root Causes and Preventability.} Backend divergence stems from preventable defaults and fundamental hardware optimizations. While users can standardize overrides like repetition penalties, optimization-induced variance (\eg non-associative reductions, hardware race conditions) is deeply tied to backend-specific implementation choices and optimized execution paths. Completely mitigating this variance is impractical, as these optimizations are required for high-throughput serving. 

\textbf{Implications for Leaderboards.} Backend choice alone can shift performance by up to $16.6$ percentage points. While the most extreme deviations stem from correctable hidden defaults, researchers are largely unaware of them. Even after correcting these, optimization-induced numerical drift still shifts scores by margins larger than the fractions of a percent frequently used to claim SOTA. This complicates comparative benchmarking, particularly when performance margins are small or inference stacks differ, as apparent gains may reflect backend artifacts rather than architectural superiority.

\textbf{No Universally Preferred Backend.} Backend-induced variance does not produce a consistent performance ordering: although transformers has the best aggregate mean rank, the preferred backend varies across models and tasks (Appendix~\ref{app:best_backend}). This makes backend choice an additional experimental variable and reinforces the need to fix and report the inference stack before evaluation.

\textbf{Security and Robustness.} Backend variance also introduces a critical ``deployment gap.'' An identical model's vulnerability to jailbreaks fluctuates by 8.9 percentage points simply by switching the backend, highlighting backend selection as a previously unrecognized security variable.

\subsection{Recommendations}
Our findings suggest that inference reproducibility is a systemic issue and thus we propose the following recommendations for researchers, evaluators, and system developers.

\noindent\textbf{Researchers and Practitioners.} \textbf{\emph{Category-Specific Reporting:}} Reporting requirements should reflect the level of control over the inference stack: \textsc{C1} \emph{Self-Hosted:} Report the inference engine and version, exact model artifact and precision, hardware, decoding parameters, prompt/chat template handling, and relevant optimization settings. Where possible, release environment specifications such as a container or pinned dependency file. \textsc{C2} \emph{Managed:} Report the provider, exact model identifier, evaluation date, all API parameters, and any model revision or system fingerprints exposed by the service. Because hidden components may change without user control, important evaluations should be repeated and interpreted as time dependent.
\textsc{C3} \emph{Routers:} Report the router and API/version, routing policy, selected provider and model for each request where available, and whether retries or fallbacks were enabled. Exact request parameters and relevant response metadata should be retained. When neither the route nor underlying model revision can be fixed, results should be described as time-dependent.
\emph{\textbf{Ensure Fair Comparisons:}}
If a SOTA method's backend is undocumented, rerun comparisons using the same backend.
\emph{\textbf{Account for Non-Determinism:}}
Repeated runs can quantify within-backend variability but cannot eliminate systematic differences between engines. Researchers should report these sources of variance separately.
\textbf{\emph{Report Per-Sample Scores:}} Alongside aggregate metrics, provide per-sample scores to enable paired statistical tests that quantify uncertainty in benchmark differences~\cite{ICLR2026_70de9e39,miller2024addingerrorbarsevals}.

\noindent\textbf{Benchmarks and Leaderboards.}
To ensure meaningful comparisons and trustworthy rankings, evaluation platforms must adopt stricter reporting and measurement practices.
\emph{\textbf{Standardize the Inference Stack:}}
Leaderboards must explicitly state the engine and configurations used, as backend-induced variance can exceed margins separating SOTA models.
\emph{\textbf{Quantify Backend Sensitivity:}}
Where feasible, evaluate using both a reference implementation and a high-throughput engine, and report the score difference.

\noindent\textbf{Inference Engine Developers and Providers.}
System-level improvements are necessary to enable reproducibility guarantees and better transparency for downstream users.
\emph{\textbf{Expose System Fingerprints:}}
For API-based inference, fixing a random seed is insufficient for reproducibility. Providers should include a system fingerprint in response metadata to allow users to track backend changes over time.
\emph{\textbf{Deterministic Modes:}}
We encourage developers to implement ``strict reproducibility'' flags. While high-performance non-deterministic kernels are essential for production, a slower, deterministic execution path is necessary for scientific debugging and validation.

\subsection{Limitations}\label{subsec:limitations}
Our controlled setup isolates Category 1 self-hosted backends by fixing model artifacts, hardware, prompts, decoding, and software versions. It therefore does not capture additional variability from production concurrency, batching regimes, hardware changes, or Category 2/3 systems, where model revisions, routing, retries, fallbacks, and provider updates may be hidden. Our primary focus is behavioral and evaluation reproducibility; Appendix~\ref{app:efficiency_metrics} reports complementary GSM8K throughput, TTFT, and inter-token latency, but we do not benchmark serving efficiency across workloads or deployment configurations.

We evaluate open-weight transformer text-generation models from 4B to 30B parameters. Our results apply to this model scale. We additionally do not evaluate multimodal, diffusion, or audio models, whose inference stacks may introduce different sources of variance.

We use greedy decoding to reduce sampling randomness, although it is not optimal for every model and can induce repetition in some reasoning models (\eg DeepSeek R1 Distill Qwen 7B). However, our goal is relative backend differences rather than optimized absolute scores. Appendix~\ref{app:sampling} shows that substantial backend differences persist at $T=0.7$. Finally, ablating every custom kernel is infeasible, and fully disabling these features to achieve perfect determinism is often impossible without abandoning the engine entirely.

%% file: content/10_conclusion.tex
\section{Conclusion}

Our results establish inference backends as an important variable in LLM evaluation. Across models, benchmarks, and metrics, we find that backend-induced numerical differences can propagate into divergent generations, altered decision boundaries, and benchmark score shifts large enough to affect model rankings. These effects are particularly pronounced for reasoning-oriented models, where early token-level divergence can fundamentally reshape the generated chain of thought. We demonstrate that this behavior is driven by a combination of hidden engine defaults and compounding numerical drift caused by essential high-performance features. At the same time, our large-scale survey of recent ML publications shows that this source of variability is rarely documented, despite the widespread use of optimized inference engines in both research and deployment.\\While the community has developed careful conventions around datasets, decoding strategies, and random seeds, the inference stack itself remains largely invisible, even though it can dominate other sources of variance. We therefore recommend reporting backend configurations, repeating evaluations, and supporting deterministic modes for scientific validation. 

%% file: content/12_appendix.tex

\section{Backend Selection and Versions}\label{app:backend_versions}

\textbf{Backend Selection.} Our landscape survey identifies many inference engines, but a controlled comparison requires a consistent setup. We therefore selected the five self-hosted engines used in our evaluation based on adoption and experimental compatibility.

We restrict the controlled comparison to Category 1 engines because they let us hold model artifacts, prompts, decoding parameters, hardware, and software versions fixed. Managed platforms may hide or update the backend, model revision, batching strategy, or hardware; routers additionally introduce routing, parameter translation, retries, and fallbacks. Evaluations in those settings therefore characterize the full hosted inference path rather than an isolated backend.

For adoption, we prioritized widely used self-hosted engines identified in Section~\ref{sec:landscape_survey}. For compatibility, each engine had to support our models, hardware, tasks, and unified software environment. This yielded vLLM~\cite{10.1145/3600006.3613165}, SGLang~\cite{zheng2024sglangefficientexecutionstructured}, llama.cpp~\cite{llama_cpp}, LMDeploy~\cite{2023lmdeploy, zhang2025efficient}, and Ollama~\cite{ollama}, with Hugging Face transformers~\cite{wolf-etal-2020-transformers} as the reference implementation.

These requirements excluded some otherwise widely adopted inference engines. For example, TensorRT-LLM~\cite{tensorrt_llm} introduced dependency conflicts with other evaluated backends, including vLLM, and could not be incorporated without changing the unified environment. Its exclusion therefore reflects experimental compatibility rather than a lack of relevance or adoption.

\textbf{Software versions.} Table~\ref{tab:backend_versions} lists the inference-backend and reference-library versions used in the controlled experiments. We enforced these fixed versions across all evaluation runs to ensure that any observed numerical variance is strictly attributable to the architectural differences between the backends.

\begin{table*}[ht]
\centering
\caption{\textbf{Software Versioning.} Specific software versions for the inference backends and reference libraries used in our evaluation.}
\label{tab:backend_versions}
\begin{tabular}{lc}
\toprule
\textbf{Library / Backend} & \textbf{Version} \\
\midrule
transformers & 4.57.0 \\
vLLM & 0.10.2 \\
SGLang & 0.5.2 \\
LMDeploy & 0.10.1 \\
llama\_cpp\_python & 0.3.16 \\
ollama & 0.13.5 (python 0.6.1) \\
\bottomrule
\end{tabular}
\end{table*}

\section{Inference Efficiency}\label{app:efficiency_metrics}
Optimized inference engines are primarily adopted for their serving efficiency. To contextualize the practical trade-off underlying our study, we measure throughput, time-to-first-token (TTFT), P95 TTFT, and inter-token latency (ITL) for every model-backend combination on GSM8K under the same setting as our controlled evaluation. These results provide complementary context rather than a comprehensive serving benchmark, as efficiency depends on workload characteristics, batching, concurrency, sequence length, and engine-specific tuning.

\begin{table}[!h]
    \centering
    \caption{
        \textbf{Inference Efficiency on GSM8K.}
        Throughput is reported in generated tokens/s; TTFT and ITL are
        reported in seconds. All measurements use the same setting as the controlled evaluation. Best values for each model are shown in bold.
    }
    \label{tab:efficiency}
    \resizebox{\textwidth}{!}{%
    \begin{tabular}{llrrrr}
        \toprule
        Model & Backend
        & Throughput $\uparrow$
        & TTFT $\downarrow$
        & P95 TTFT $\downarrow$
        & ITL $\downarrow$ \\
        \midrule

        Qwen3 4B
        & transformers & 58.146  & 0.025 & 0.030 & 0.015 \\
        & llama.cpp     & 123.694 & 0.021 & 0.023 & 0.007 \\
        & LMDeploy      & \textbf{153.517} & \textbf{0.015} & 0.019 & \textbf{0.005} \\
        & Ollama        & 136.685 & 0.088 & 0.090 & 0.006 \\
        & SGLang        & 147.788 & 0.019 & 0.023 & \textbf{0.005} \\
        & vLLM          & 140.209 & \textbf{0.015} & \textbf{0.015} & 0.006 \\
        \midrule

        Llama 3.1 8B
        & transformers & 76.259  & 0.027 & 0.030 & 0.012 \\
        & llama.cpp     & 102.361 & 0.026 & 0.027 & 0.009 \\
        & LMDeploy      & \textbf{112.834} & \textbf{0.021} & \textbf{0.022} & \textbf{0.008} \\
        & Ollama        & 100.140 & 0.112 & 0.115 & 0.009 \\
        & SGLang        & 109.071 & 0.023 & 0.025 & \textbf{0.008} \\
        & vLLM          & 106.996 & \textbf{0.021} & \textbf{0.022} & 0.009 \\
        \midrule

        Qwen3 30B
        & transformers & 12.255  & 0.548 & 0.586 & 0.070 \\
        & llama.cpp     & 93.747  & 0.159 & 0.173 & 0.008 \\
        & LMDeploy      & 131.651 & \textbf{0.032} & \textbf{0.036} & \textbf{0.006} \\
        & Ollama        & 100.993 & 0.121 & 0.163 & 0.008 \\
        & SGLang        & \textbf{137.455} & 0.038 & 0.043 & \textbf{0.006} \\
        & vLLM          & 126.893 & 0.035 & 0.042 & \textbf{0.006} \\
        \midrule

        DeepSeek R1 Distill Qwen 7B
        & transformers & 72.400  & 0.024 & 0.030 & 0.013 \\
        & llama.cpp     & 92.430  & 0.025 & 0.026 & 0.010 \\
        & LMDeploy      & \textbf{114.157} & \textbf{0.019} & 0.023 & \textbf{0.008} \\
        & Ollama        & 103.984 & 0.095 & 0.098 & \textbf{0.008} \\
        & SGLang        & 108.928 & 0.021 & 0.026 & \textbf{0.008} \\
        & vLLM          & 106.494 & \textbf{0.019} & \textbf{0.019} & 0.009 \\
        \midrule

        Qwen3 Thinking 30B
        & transformers & 12.827  & 0.552 & 0.589 & 0.070 \\
        & llama.cpp     & 103.055 & 0.165 & 0.180 & 0.009 \\
        & LMDeploy      & 139.336 & \textbf{0.036} & \textbf{0.039} & \textbf{0.006} \\
        & Ollama        & 111.773 & 0.124 & 0.164 & 0.008 \\
        & SGLang        & \textbf{152.070} & 0.042 & 0.046 & \textbf{0.006} \\
        & vLLM          & 136.762 & 0.039 & 0.046 & \textbf{0.006} \\
        \bottomrule
    \end{tabular}
    }
\end{table}

As shown in Table~\ref{tab:efficiency}, the optimized backends provide substantial efficiency gains over transformers, particularly for the 30B models, while no backend dominates across all models and metrics. For example, SGLang achieves the highest throughput for both 30B models, whereas LMDeploy achieves the lowest mean TTFT. These results illustrate the practical incentive for optimized inference engines that motivates our study, while our main experiments examine whether these gains come with changes in model behavior.

\newpage

\section{Universal Best Backend Analysis}\label{app:best_backend}
The performance differences in Table~\ref{tab:comparison_experiment} raise the question of whether one inference backend systematically achieves better benchmark results. Because the benchmarks use different metrics and scales, we rank the six implementations within each of the 20 model-benchmark combinations and average their ranks. Rank 1 denotes the highest score; ties receive their average rank.
\begin{table}[!h]
    \centering
    \caption{\textbf{Aggregate Backend Rankings.}
    Mean rank across the 20 model-benchmark combinations in Table~\ref{tab:comparison_experiment}
    (lower is better). Engines are ranked within each combination according to their mean benchmark score; ties receive average ranks.
    Top finishes report sole and tied highest-scoring combinations separately.}
    \label{tab:backend_rank_summary}
    \small
    \begin{tabular}{lcc}
        \toprule
        \textbf{Backend} & \textbf{Mean Rank $\downarrow$}
        & \textbf{Top Finishes (Sole / Tied)} \\
        \midrule
        transformers & \textbf{2.83} & \textbf{7 / 1} \\
        vLLM         & 3.33 & 3 / 2 \\
        LMDeploy     & 3.55 & 2 / 2 \\
        SGLang       & 3.63 & 2 / 1 \\
        llama.cpp    & 3.68 & 1 / 0 \\
        Ollama       & 4.00 & 2 / 1 \\
        \bottomrule
    \end{tabular}
\end{table}\\
Table~\ref{tab:backend_rank_summary} shows that transformers obtains the best aggregate mean rank (2.83), but is the sole top-performing backend in only 7 of 20 combinations. Moreover, paired row-bootstrap analysis does not establish a better aggregate rank for transformers than any individual alternative at the 95\,\% confidence level.
\begin{table}[!h]
    \centering
    \caption{\textbf{Backend Rankings Depend on the Evaluation Setting.}
    Best mean-rank backend when aggregating separately within each model
    (across four benchmarks) and benchmark (across five models).}
    \label{tab:backend_rank_by_setting}
    \small
    \begin{tabular}{lcc}
        \toprule
        \textbf{Subset} & \textbf{Best Backend}
        & \textbf{Mean Rank $\downarrow$} \\
        \midrule
        \multicolumn{3}{l}{\textit{By model}} \\
        Qwen3 4B                         & transformers & 2.50 \\
        Llama 3.1 8B                     & LMDeploy     & 2.13 \\
        Qwen3 30B                        & Ollama       & 1.88 \\
        DeepSeek R1 Distill Qwen 7B      & vLLM         & 2.00 \\
        Qwen3 Thinking 30B               & transformers & 1.50 \\
        \midrule
        \multicolumn{3}{l}{\textit{By benchmark}} \\
        GPQA                             & transformers & 2.70 \\
        GSM8K                            & transformers & 1.90 \\
        SimpleQA                         & transformers & 3.00 \\
        LiveCodeBench                    & vLLM         & 1.80 \\
        \bottomrule
    \end{tabular}
\end{table}\\
The preferred backend also depends on the evaluated setting (Table~\ref{tab:backend_rank_by_setting}). Four different backends achieve the best average rank for at least one model, and vLLM ranks first on LiveCodeBench. The aggregate result is also sensitive to the specific evaluation configuration: the leader changes from transformers to vLLM when either GSM8K or Qwen3 Thinking 30B is omitted.\\
Thus, although transformers has the best average rank across the complete evaluation, we find no evidence for a backend that is uniformly preferable across models and tasks. Selecting a backend after observing benchmark results could improve a particular score. We therefore recommend fixing and reporting the inference stack before evaluation. Our analysis does not provide evidence that prior work intentionally exploits backend choice to inflate results.

\section{Additional Metrics}\label{app:further_metrics}
This section provides detailed visualizations and breakdowns for the evaluation metrics introduced in Section~\ref{sec:eval}. By observing these metrics across individual datasets and models, we can better understand how backend-induced variance disproportionately affects specific architectures (\eg reasoning models) and tasks.
\subsection{Disagreement Rate}
\begin{figure}[!h]
    \centering
    \begin{subfigure}[b]{0.48\textwidth}
        \centering
        \includegraphics[width=\textwidth]{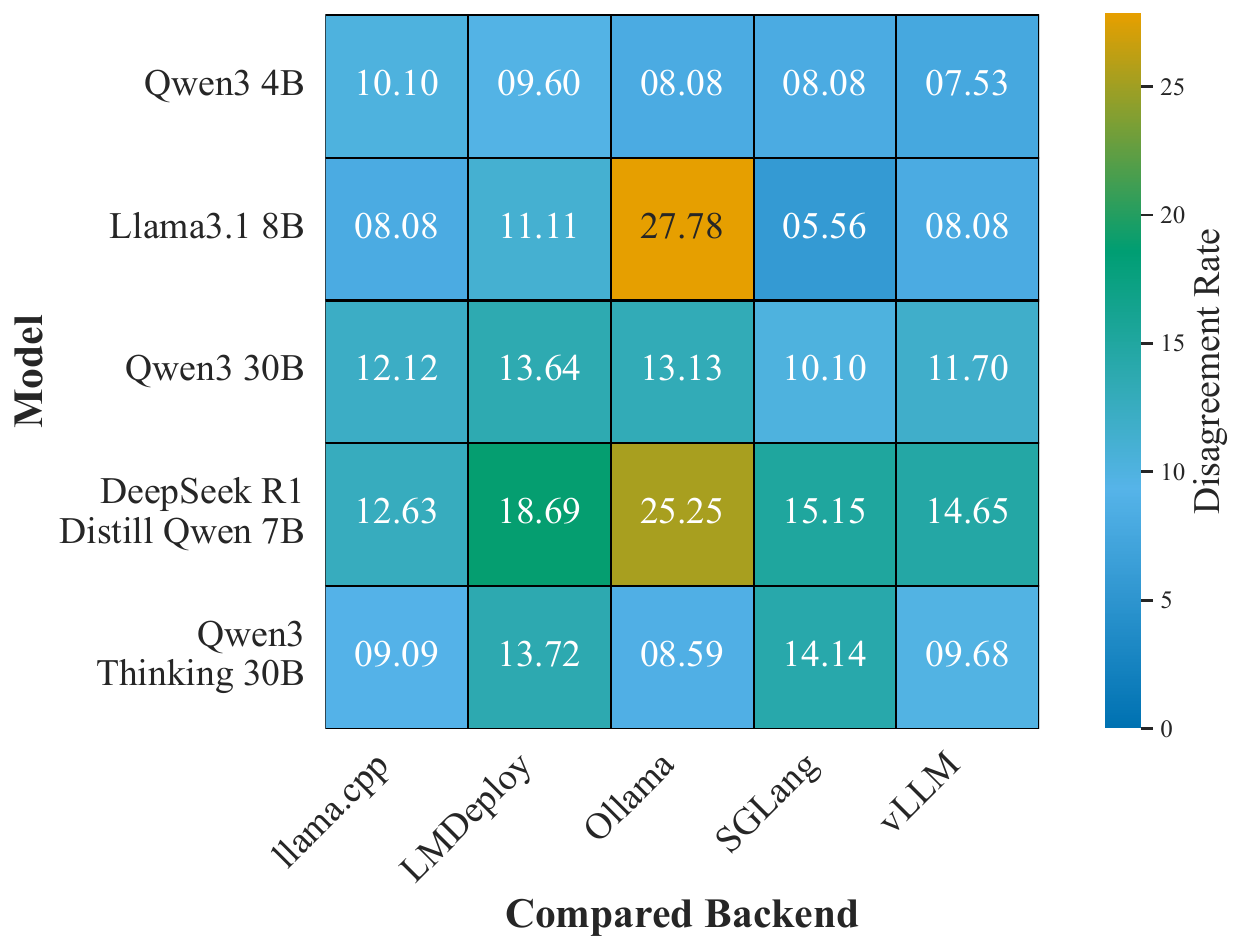}
        \caption{GPQA}
        \label{fig:disagreement_gpqa}
    \end{subfigure}
    \hfill
    \begin{subfigure}[b]{0.48\textwidth}
        \centering
        \includegraphics[width=\textwidth]{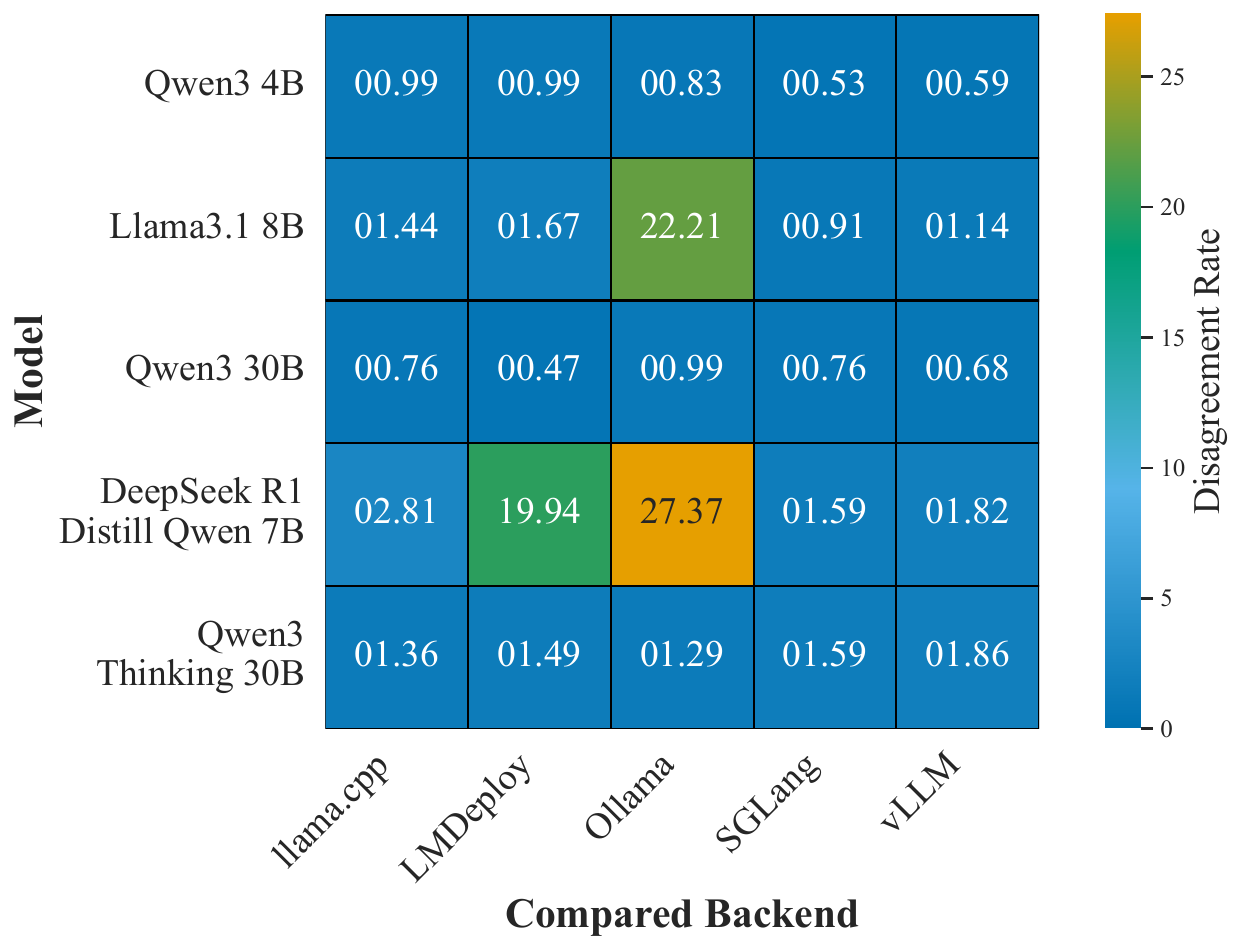}
        \caption{GSM8K}
        \label{fig:disagreement_gsm8k}
    \end{subfigure}
    
    \vspace{1.5em} 
    
    \begin{subfigure}[b]{0.48\textwidth}
        \centering
        \includegraphics[width=\textwidth]{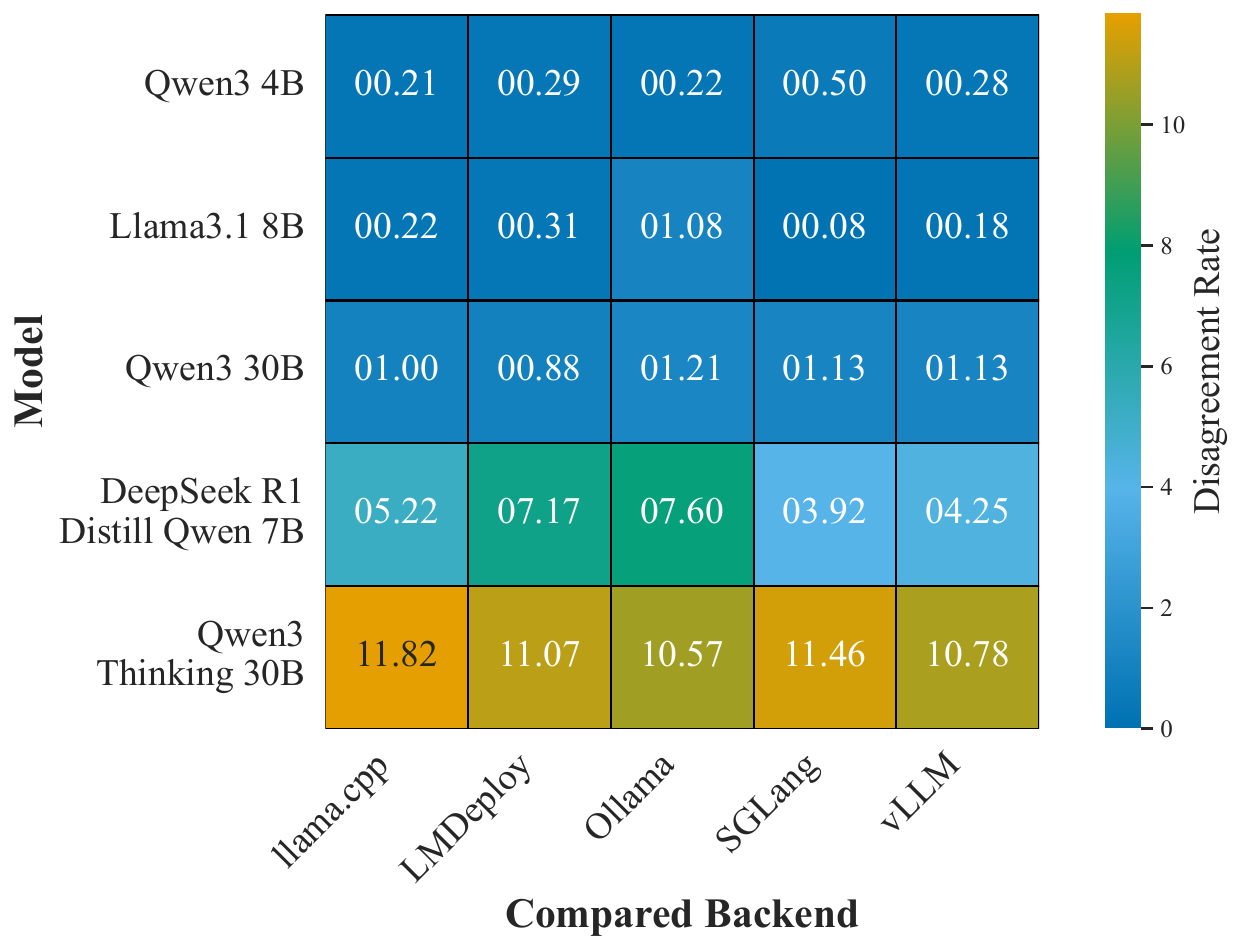}
        \caption{SimpleQA}
        \label{fig:disagreement_simpleqa}
    \end{subfigure}
    \hfill
    \begin{subfigure}[b]{0.48\textwidth}
        \centering
        \includegraphics[width=\textwidth]{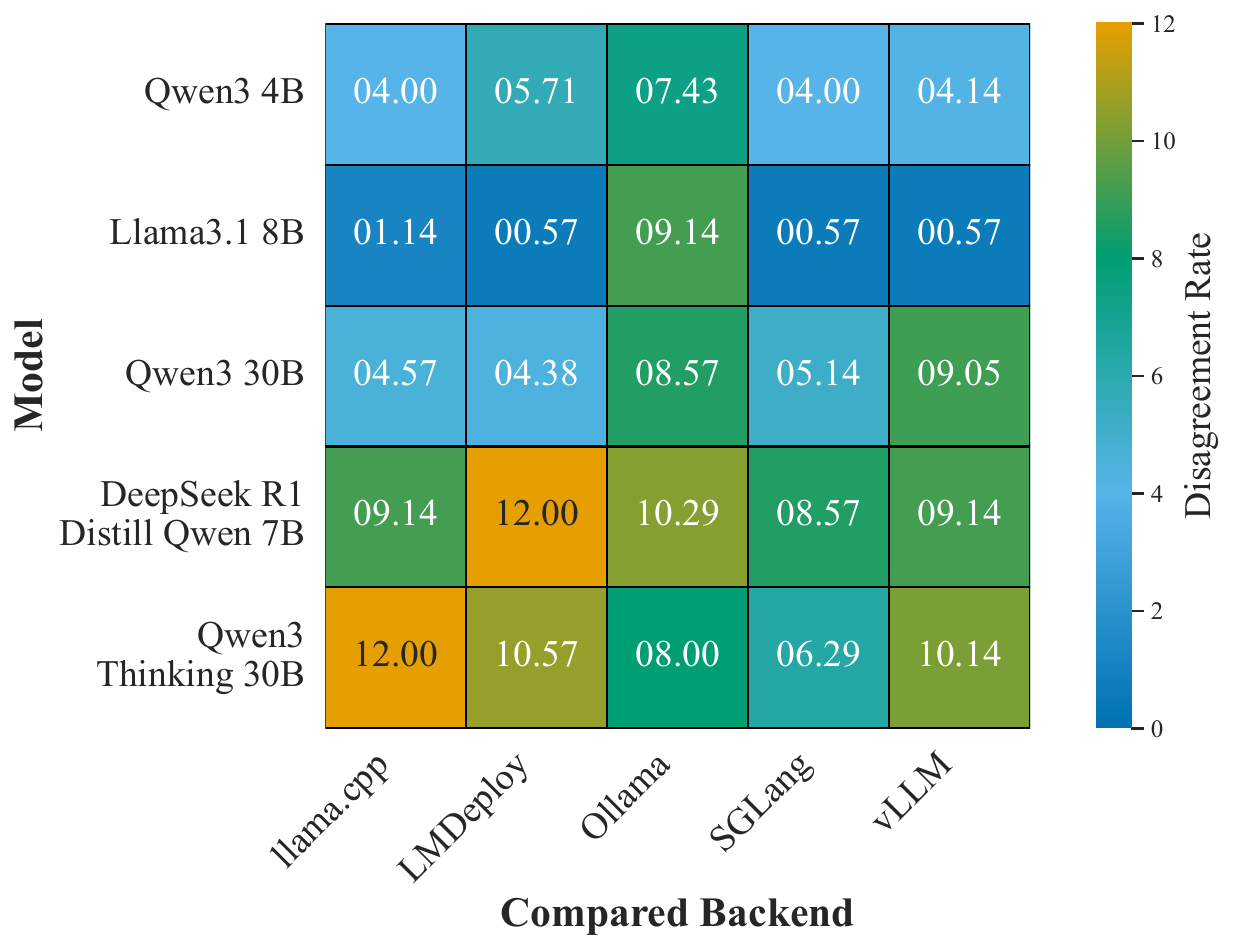}
        \caption{LiveCodeBench}
        \label{fig:disagreement_livecodebench}
    \end{subfigure}
    
    \caption{\textbf{Output Disagreement Rates.} The frequency with which each backend's prediction differs from the transformers reference implementation for the same input. Higher values indicate a larger disagreement between the two backends.}
    \label{fig:disagreement_rate}
\end{figure}

As defined in Section~\ref{subsec:eval_outcome}, the Disagreement Rate measures the absolute frequency with which a backend produces a different final prediction than the transformers reference. Figure~\ref{fig:disagreement_rate} illustrates these rates across our evaluated benchmarks. While aggregate scores (Table~\ref{tab:comparison_experiment}) might appear stable in certain configurations, the disagreement rate reveals underlying instability. Two backends can achieve the exact same overall accuracy score while correctly answering a completely different subset of questions, indicating that the backend numerical variance actively shifts the model's decision~boundary.
\label{app:diasgreement}

\newpage

\subsection{Length Error}
\label{app:length_error}
\begin{figure}[!h]
    \centering
    {
        \setlength{\fboxsep}{0pt} 
        \setlength{\fboxrule}{0.1pt} 
        \fbox{\includegraphics[width=0.7\textwidth]{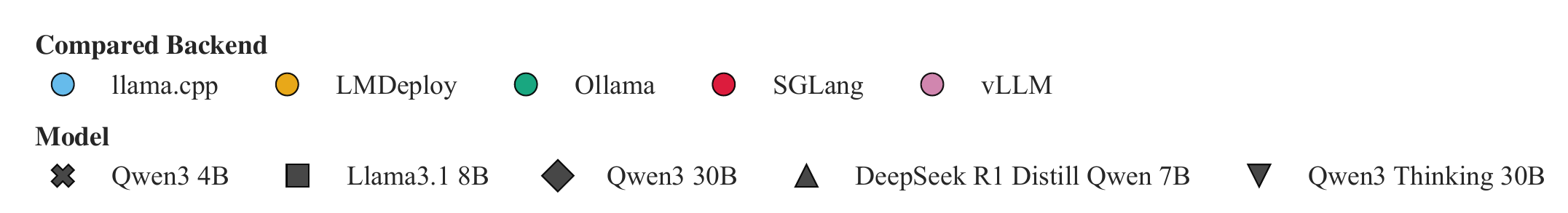}}
    }
    
    \vspace{1.5em} 
    
    \begin{subfigure}[b]{0.48\textwidth}
        \centering
        \includegraphics[width=\textwidth]{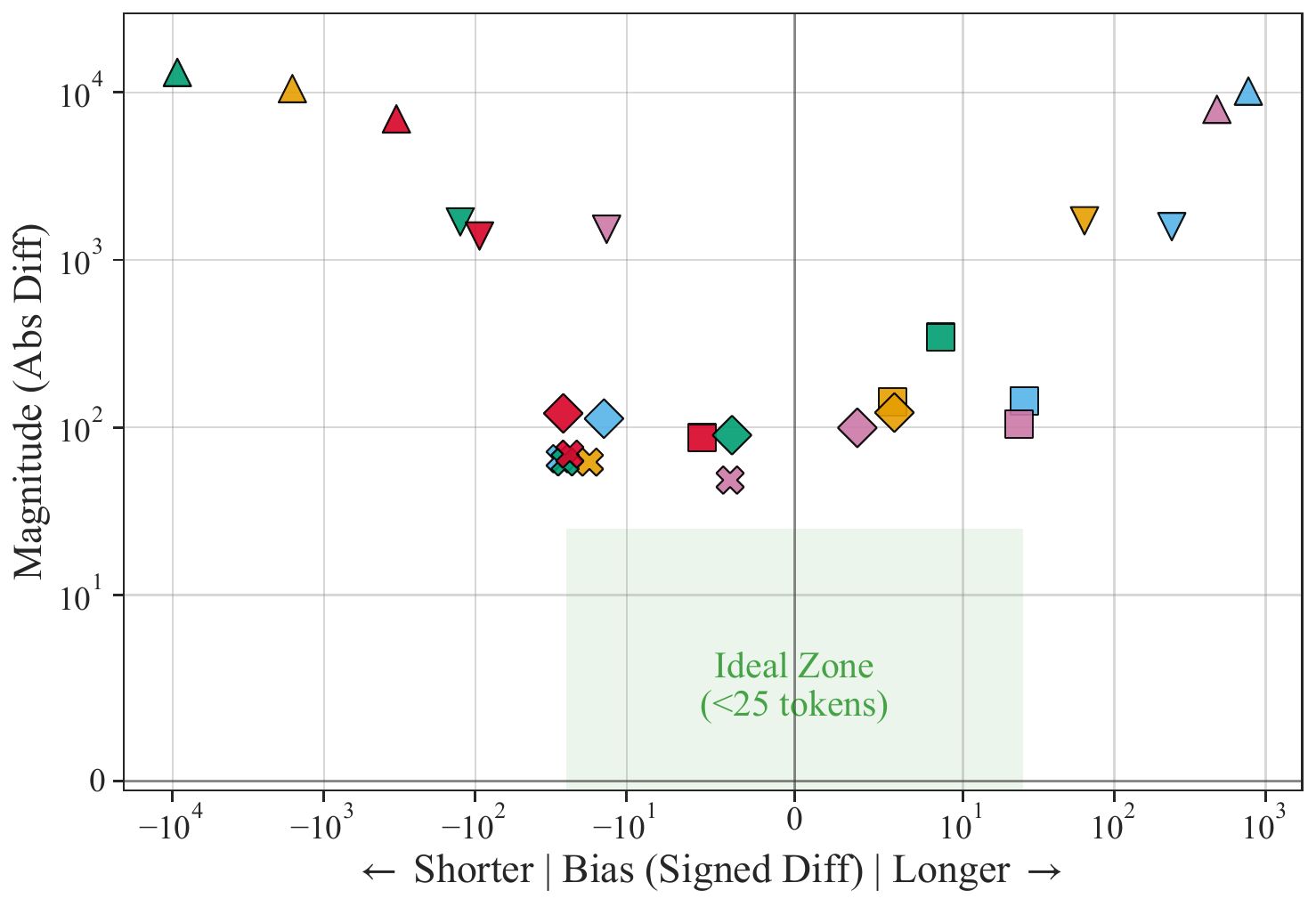}
        \caption{GPQA}
        \label{fig:length_gpqa}
    \end{subfigure}
    \hfill
    \begin{subfigure}[b]{0.48\textwidth}
        \centering
        \includegraphics[width=\textwidth]{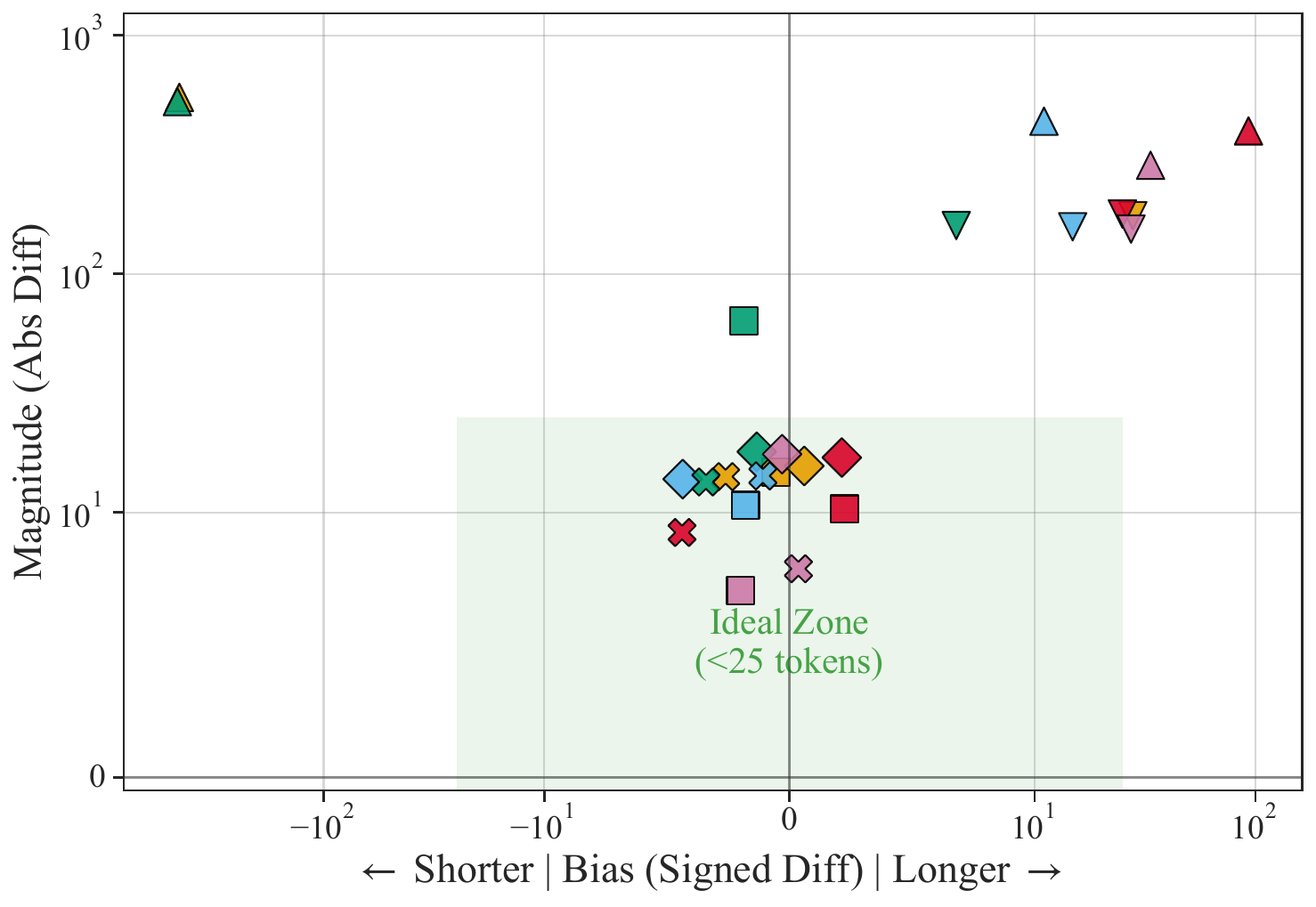}
        \caption{GSM8K}
        \label{fig:length_gsm8k}
    \end{subfigure}
    
    \vspace{1.5em} 
    
    \begin{subfigure}[b]{0.48\textwidth}
        \centering
        \includegraphics[width=\textwidth]{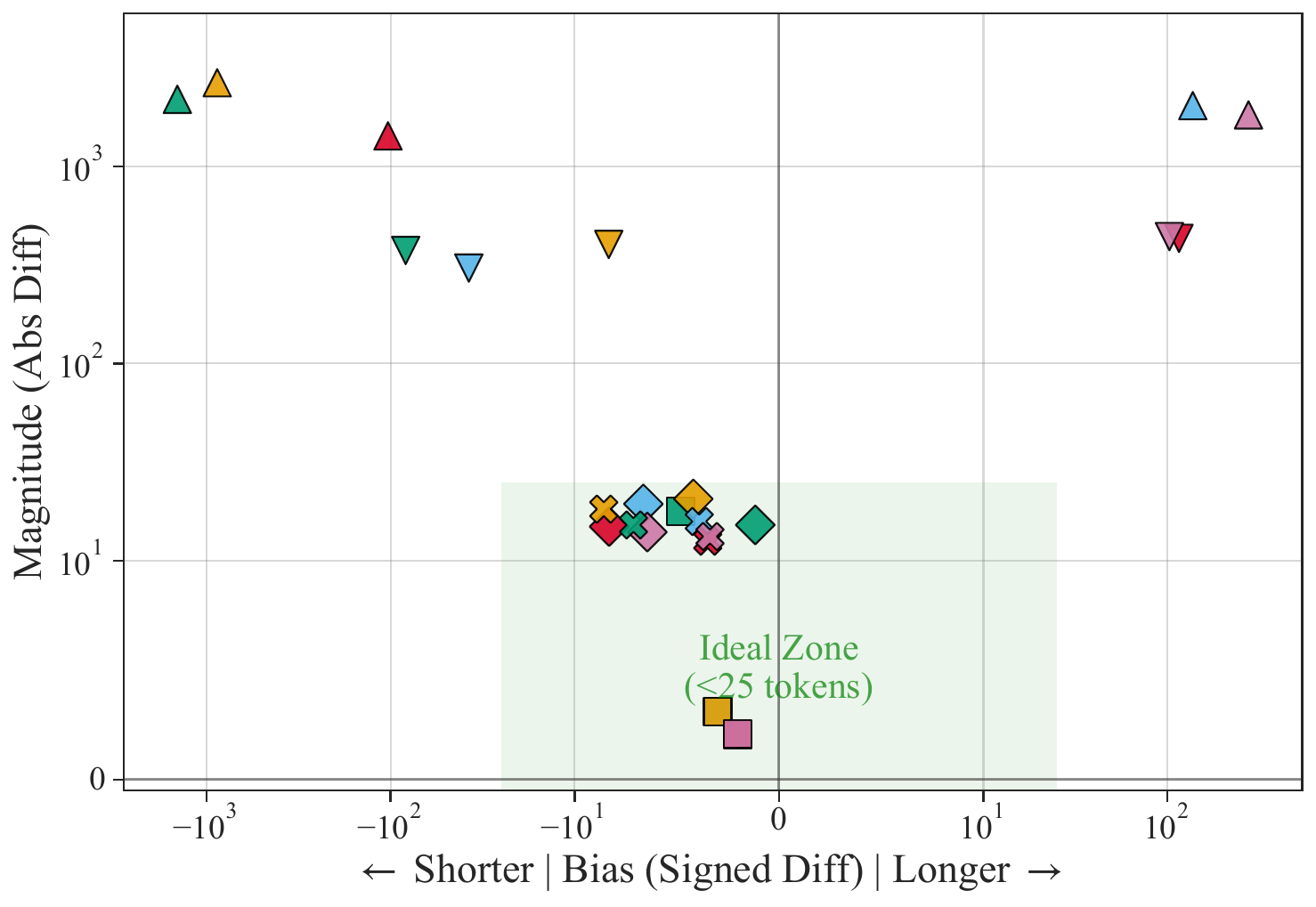}
        \caption{SimpleQA}
        \label{fig:length_simpleqa}
    \end{subfigure}
    \hfill
    \begin{subfigure}[b]{0.48\textwidth}
        \centering
        \includegraphics[width=\textwidth]{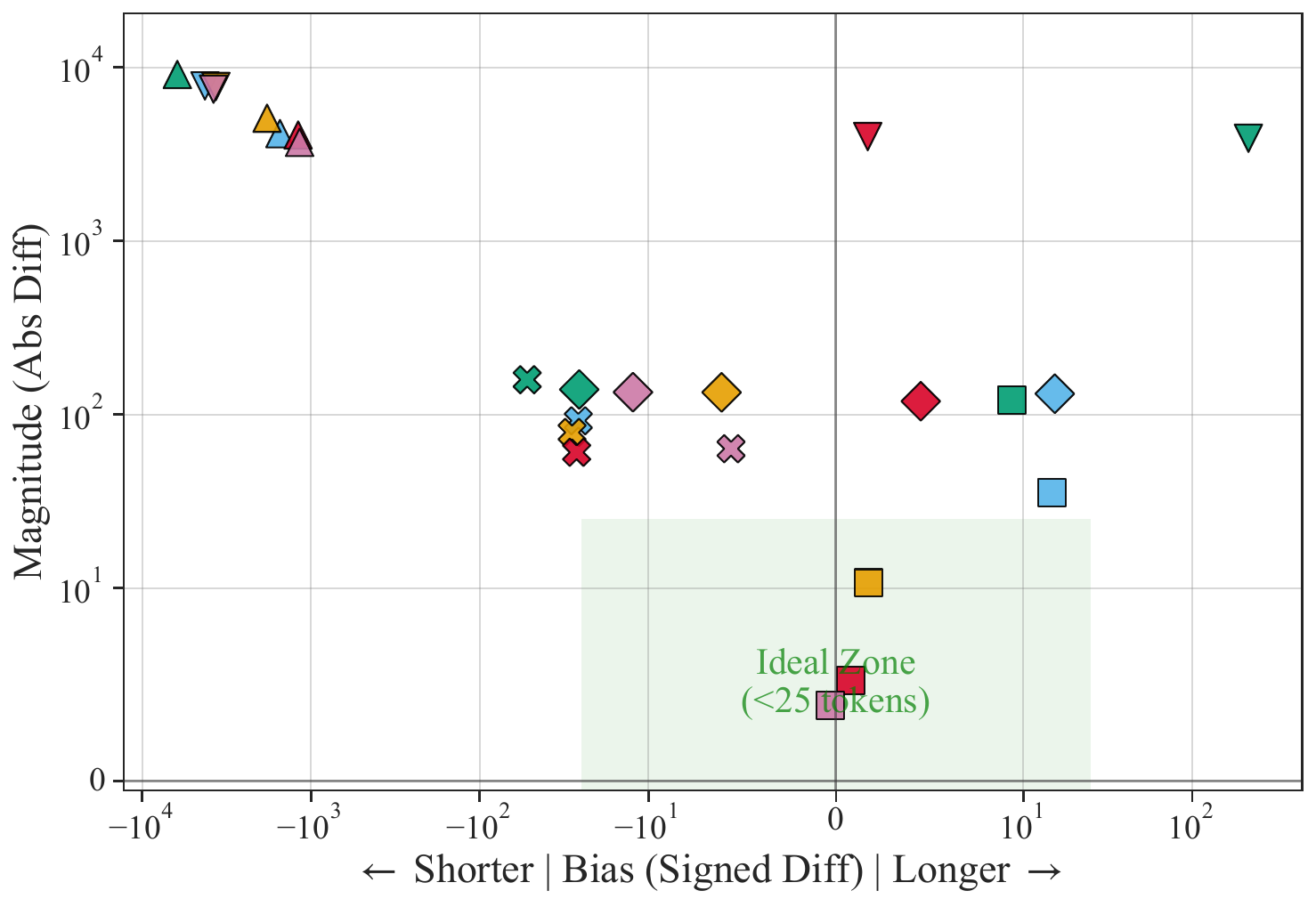}
        \caption{LiveCodeBench}
        \label{fig:length_livecodebench}
    \end{subfigure}
    
    \caption{\textbf{Analysis of Output Length Consistency against the transformers Reference.} This scatter plot visualizes the deviation in generation length for various backends. The \textbf{X-axis (Bias)} represents the average \textit{Signed Difference}, where negative values indicate the backend generated fewer tokens than the reference (shorter), and positive values indicate more tokens (longer). The \textbf{Y-axis (Magnitude)} represents the average \textit{Absolute Difference}, showing the total scale of the deviation regardless of direction. The shaded green region (``Ideal Zone'') marks acceptable variance ($\pm$25 tokens), roughly equating to a 1-2 sentence difference.}
    \label{fig:length_error}
\end{figure}

Beyond final accuracy, we analyze structural deviations in the generated responses by measuring Output Length consistency (Figure~\ref{fig:length_error}). We plot both the Signed Difference (Bias) on the X-axis and the Absolute Difference (Magnitude) on the Y-axis. The signed difference reveals whether a backend has a systematic bias toward verbosity (producing consistently longer or shorter sequences), while the absolute difference captures the scale of the deviation, preventing positive and negative length differences from canceling each other out. We define an ``Ideal Zone'' of $\pm$25 tokens (roughly 1-2 sentences), where the structural integrity of the answer is largely preserved. As shown, reasoning models frequently lie outside this zone, experiencing massive shifts in generation length that fundamentally alter their chain-of-thought process.

\newpage

\subsection{Token Divergence}
\label{app:divergence}
\begin{figure}[!h]
    \centering
    \includegraphics[width=\textwidth]{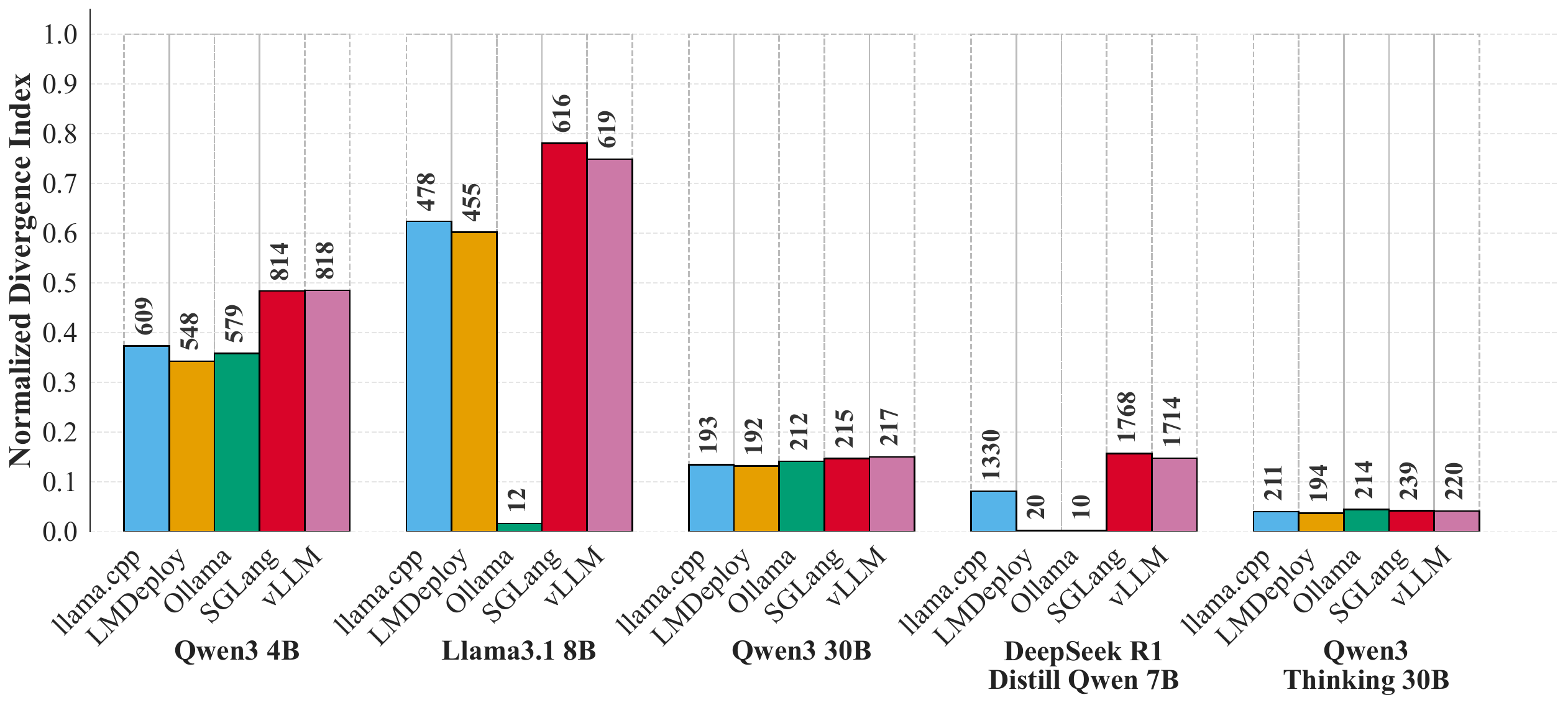}
    \caption{\textbf{Token-Level Divergence Analysis (GPQA).} Normalized divergence scores relative to the transformers reference. Larger values indicate high similarity (divergence happens late), while smaller values indicate early divergence. The labels above the bars indicate the average token position at which the generation first differs from the reference sequence.}
    \label{fig:divergence_gpqa}
    
    \vspace{3em} 
    
    \includegraphics[width=\textwidth]{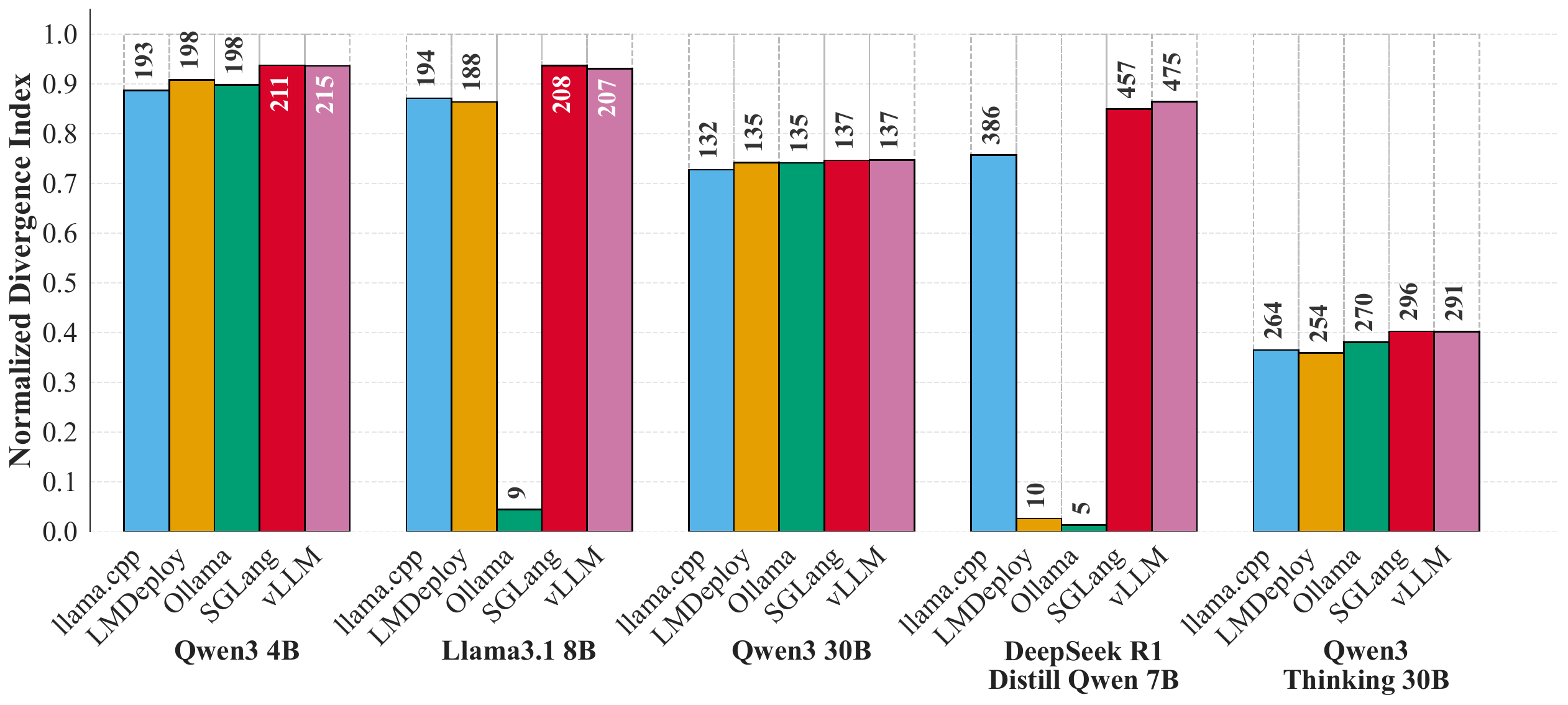}
    \caption{\textbf{Token-Level Divergence Analysis (GSM8K).} Normalized divergence scores relative to the transformers reference. Larger values indicate high similarity (divergence happens late), while smaller values indicate early divergence. The labels above the bars indicate the average token position at which the generation first differs from the reference sequence.}
    \label{fig:divergence_gsm8k}
\end{figure}

\begin{figure}[!h]
    \centering
    \includegraphics[width=\textwidth]{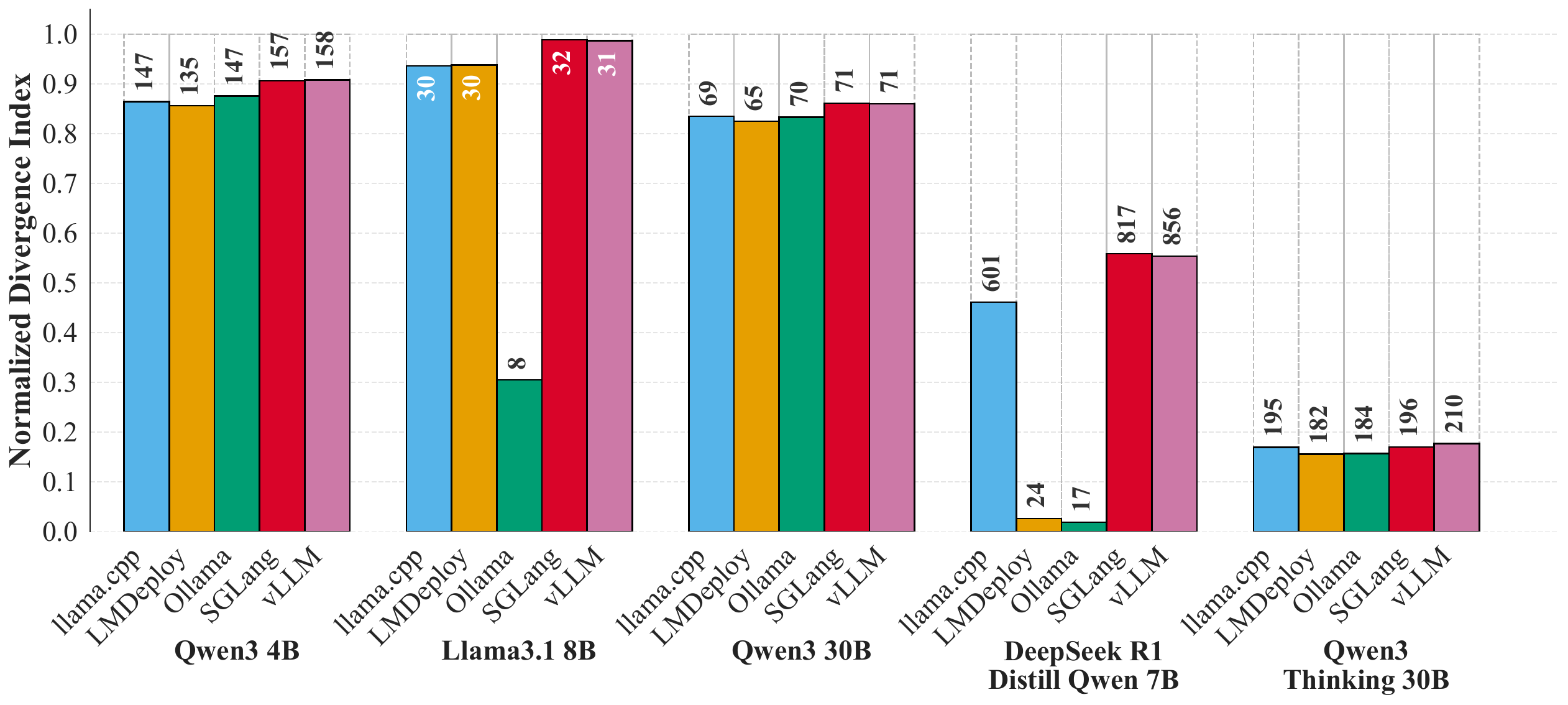}
    \caption{\textbf{Token-Level Divergence Analysis (SimpleQA).} Normalized divergence scores relative to the transformers reference. Larger values indicate high similarity (divergence happens late), while smaller values indicate early divergence. The labels above the bars indicate the average token position at which the generation first differs from the reference sequence.}
    \label{fig:divergence_simpleqa}
    
    \vspace{3em}
    
    \includegraphics[width=\textwidth]{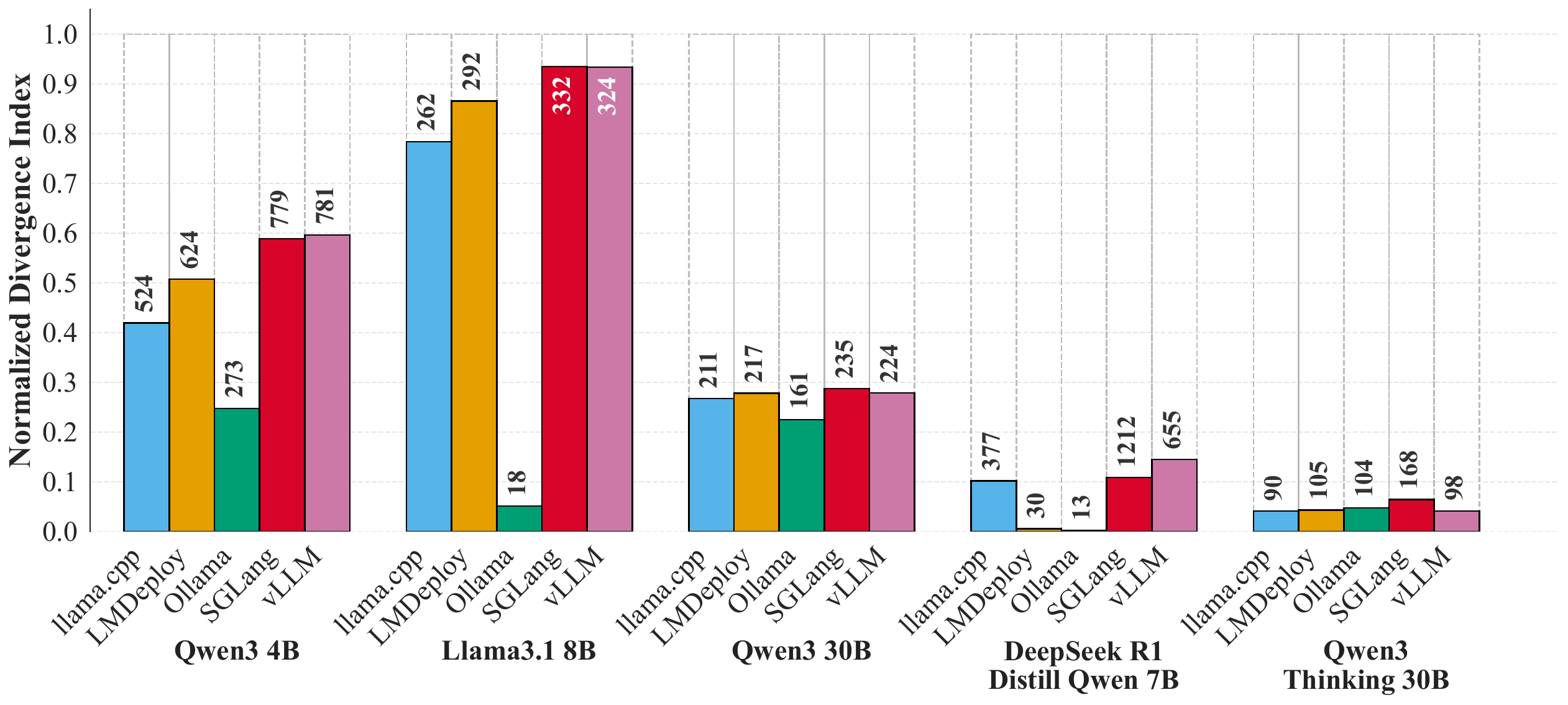}
    \caption{\textbf{Token-Level Divergence Analysis (LiveCodeBench).} Normalized divergence scores relative to the transformers reference. Larger values indicate high similarity (divergence happens late), while smaller values indicate early divergence. The labels above the bars indicate the average token position at which the generation first differs from the reference sequence.}
    \label{fig:divergence_livecodebench}
\end{figure}

To pinpoint exactly \emph{when} the generations begin to differ, we calculate the Token-Level Divergence Index. Figures~\ref{fig:divergence_gpqa} through~\ref{fig:divergence_livecodebench} visualize the Normalized Divergence Score alongside the absolute token position of the first mismatch (labels above the bars). A normalized score closer to 1.0 indicates that the generations remain identical for the majority of the sequence, whereas lower scores indicate early divergence. Our results highlight that difficult benchmarks (such as GPQA) cause models to diverge much earlier in the generation process.
\newpage

\newpage

\newpage

\newpage

\subsection{LogProb Error}
\label{app:logprob_rmse}
\begin{figure}[!h]
    \centering
    \begin{subfigure}[b]{0.48\textwidth}
        \centering
        \includegraphics[width=\textwidth]{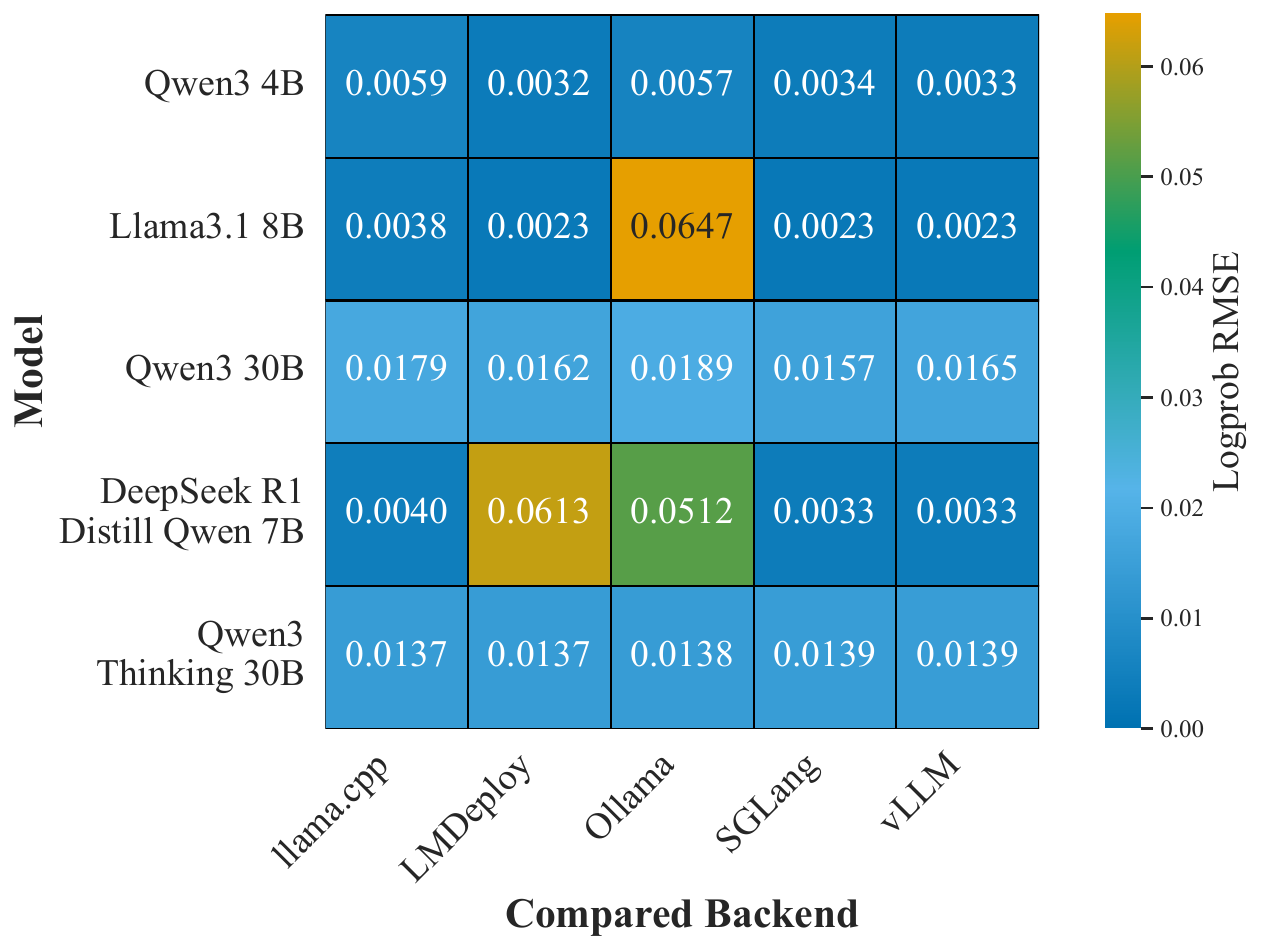}
        \caption{GPQA}
        \label{fig:logprob_gpqa}
    \end{subfigure}
    \hfill
    \begin{subfigure}[b]{0.48\textwidth}
        \centering
        \includegraphics[width=\textwidth]{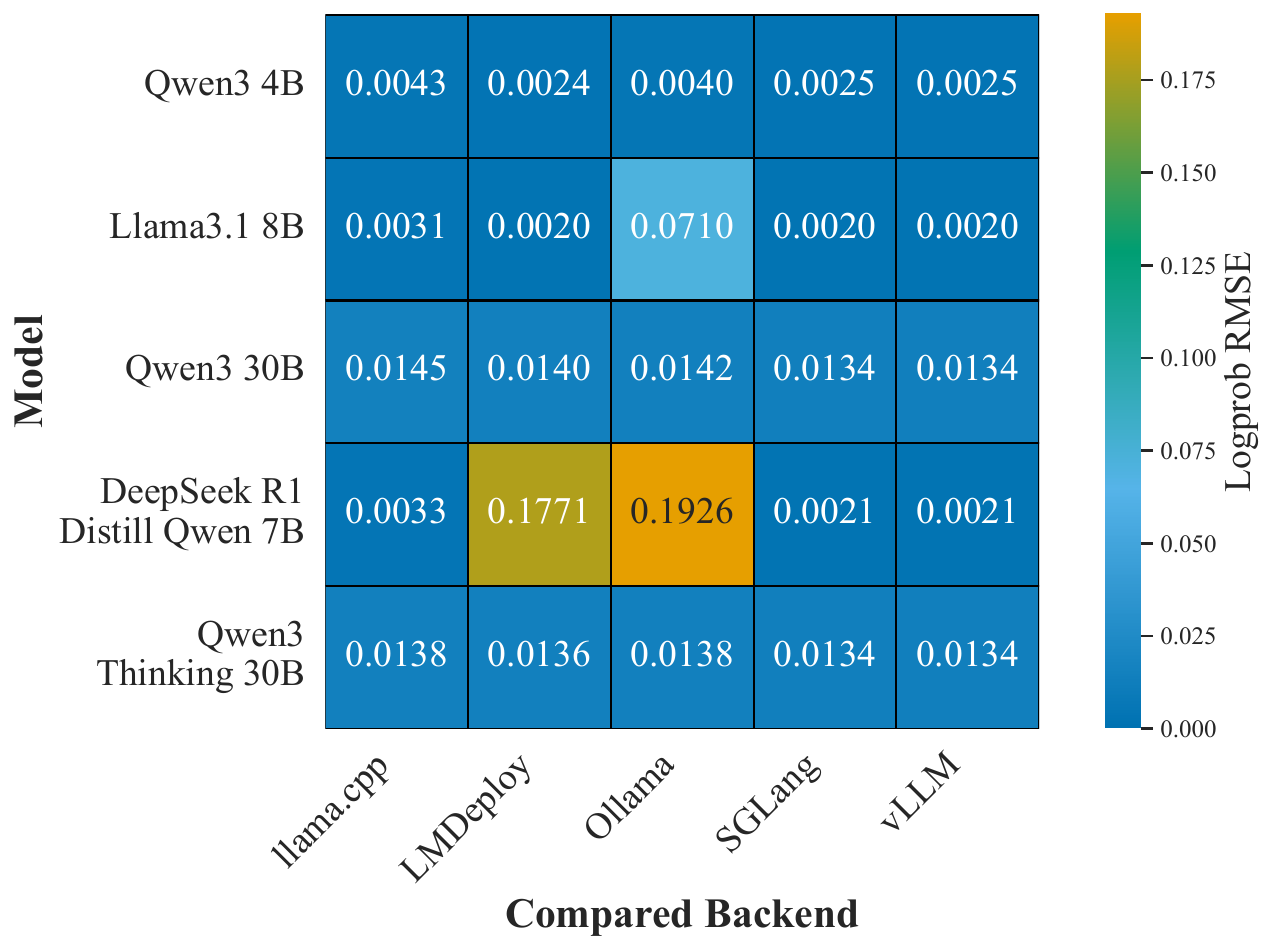}
        \caption{GSM8K}
        \label{fig:logprob_gsm8k}
    \end{subfigure}
    
    \vspace{1.5em} 
    
    \begin{subfigure}[b]{0.48\textwidth}
        \centering
        \includegraphics[width=\textwidth]{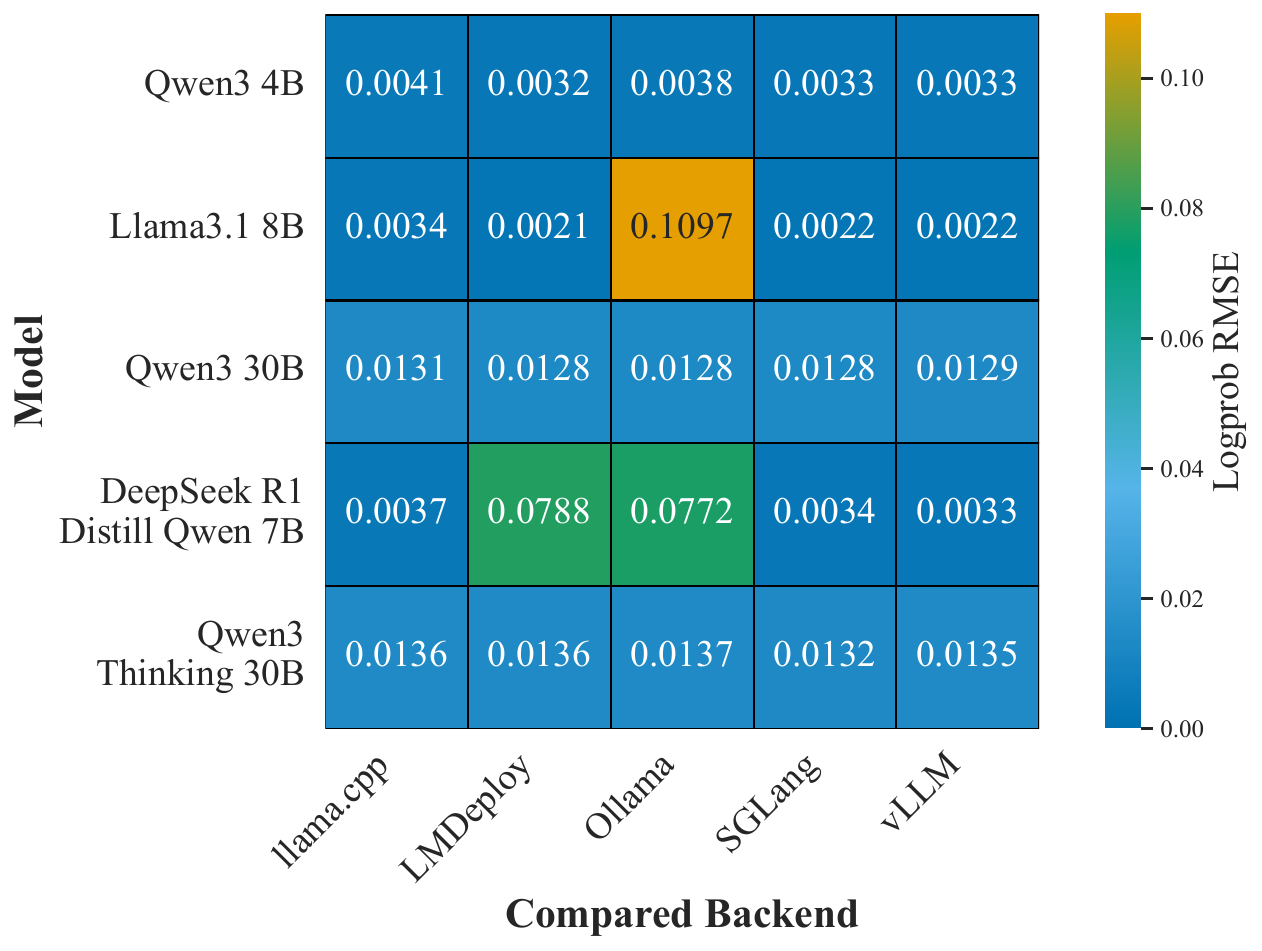}
        \caption{SimpleQA}
        \label{fig:logprob_simpleqa}
    \end{subfigure}
    \hfill
    \begin{subfigure}[b]{0.48\textwidth}
        \centering
        \includegraphics[width=\textwidth]{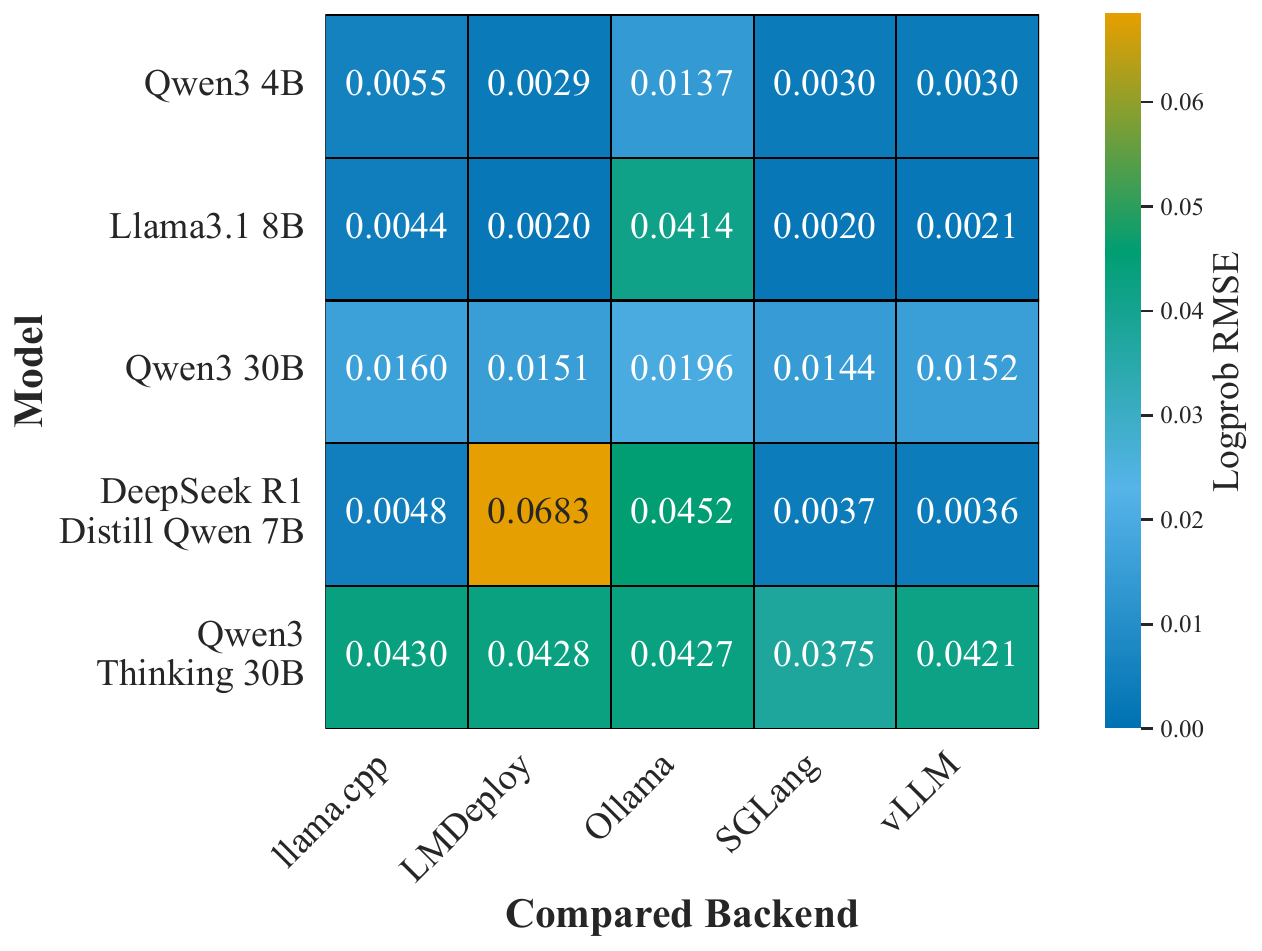}
        \caption{LiveCodeBench}
        \label{fig:logprob_livecodebench}
    \end{subfigure}
    
    \caption{\textbf{Numerical Precision (LogProb RMSE).} Root Mean Squared Error (RMSE) of the top-1 token log-probabilities compared to the transformers reference. In an ideal, deterministic setting, we expect an RMSE of exactly 0.0, indicating identical confidence in token selection. Higher values demonstrate numerical drift caused by the backend. This indicates that the underlying probability distribution is shifting, which can eventually cascade into divergent token selections.}
    \label{fig:logprob_rmse}
\end{figure}

While the previous metrics evaluate the final generated text, we also investigate the underlying floating-point stability prior to any generation mismatch. Figure~\ref{fig:logprob_rmse} presents the LogProb RMSE for the top-1 token calculated on the matching prefix (the tokens generated before the sequences diverge). These heatmaps confirm that numerical drift is present at the logit level even when the discrete greedy token selections remain identical. Reasoning models consistently exhibit higher baseline drift compared to standard models, foreshadowing their higher rates of downstream token divergence.

\newpage

\subsection{Top-5 Token Jaccard Similarity}
\begin{figure}[!h]
    \centering
    \begin{subfigure}[b]{0.48\textwidth}
        \centering
        \includegraphics[width=\textwidth]{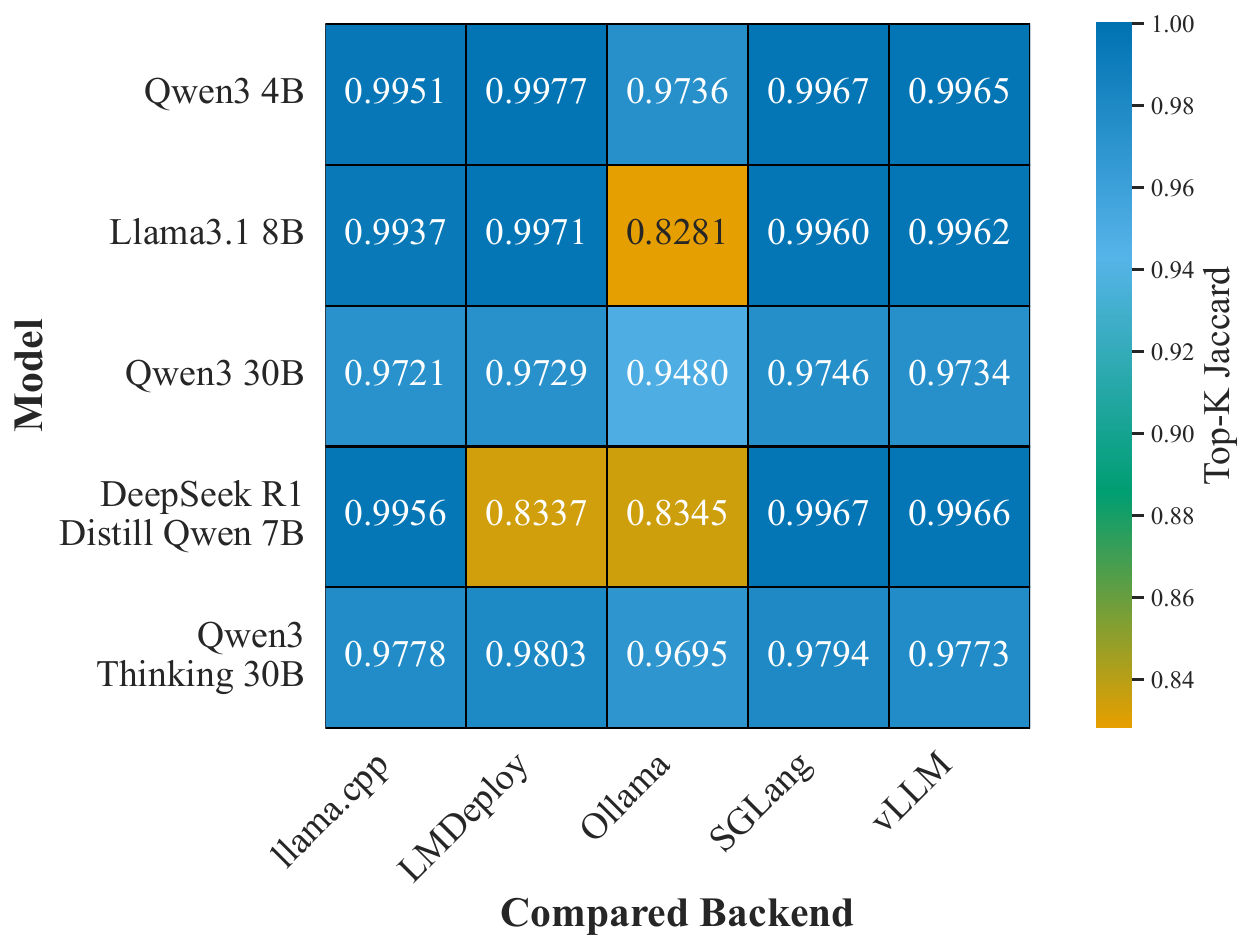}
        \caption{GPQA}
        \label{fig:jaccard_gpqa}
    \end{subfigure}
    \hfill
    \begin{subfigure}[b]{0.48\textwidth}
        \centering
        \includegraphics[width=\textwidth]{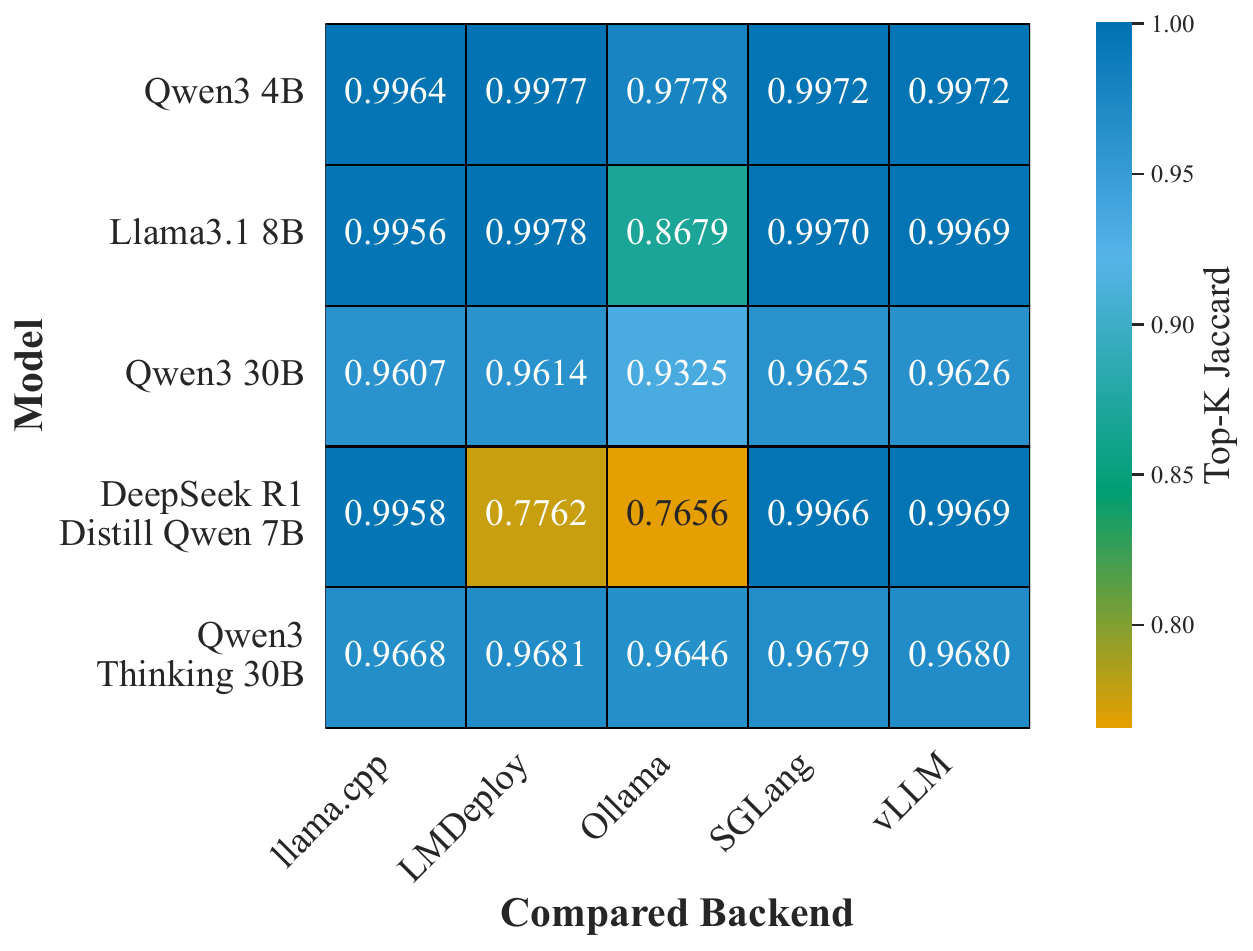}
        \caption{GSM8K}
        \label{fig:jaccard_gsm8k}
    \end{subfigure}
    
    \vspace{1.5em} 
    
    \begin{subfigure}[b]{0.48\textwidth}
        \centering
        \includegraphics[width=\textwidth]{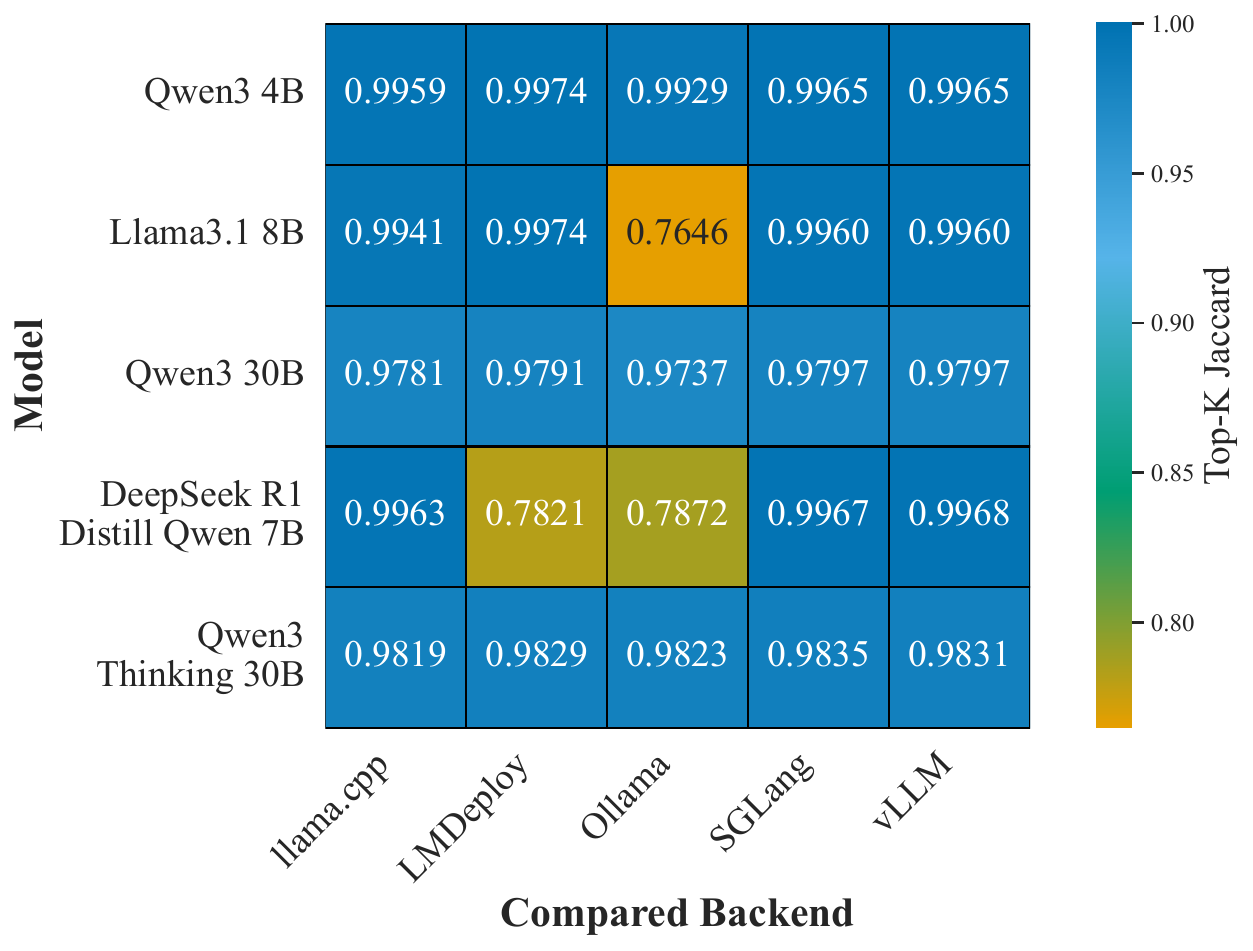}
        \caption{SimpleQA}
        \label{fig:jaccard_simpleqa}
    \end{subfigure}
    \hfill
    \begin{subfigure}[b]{0.48\textwidth}
        \centering
        \includegraphics[width=\textwidth]{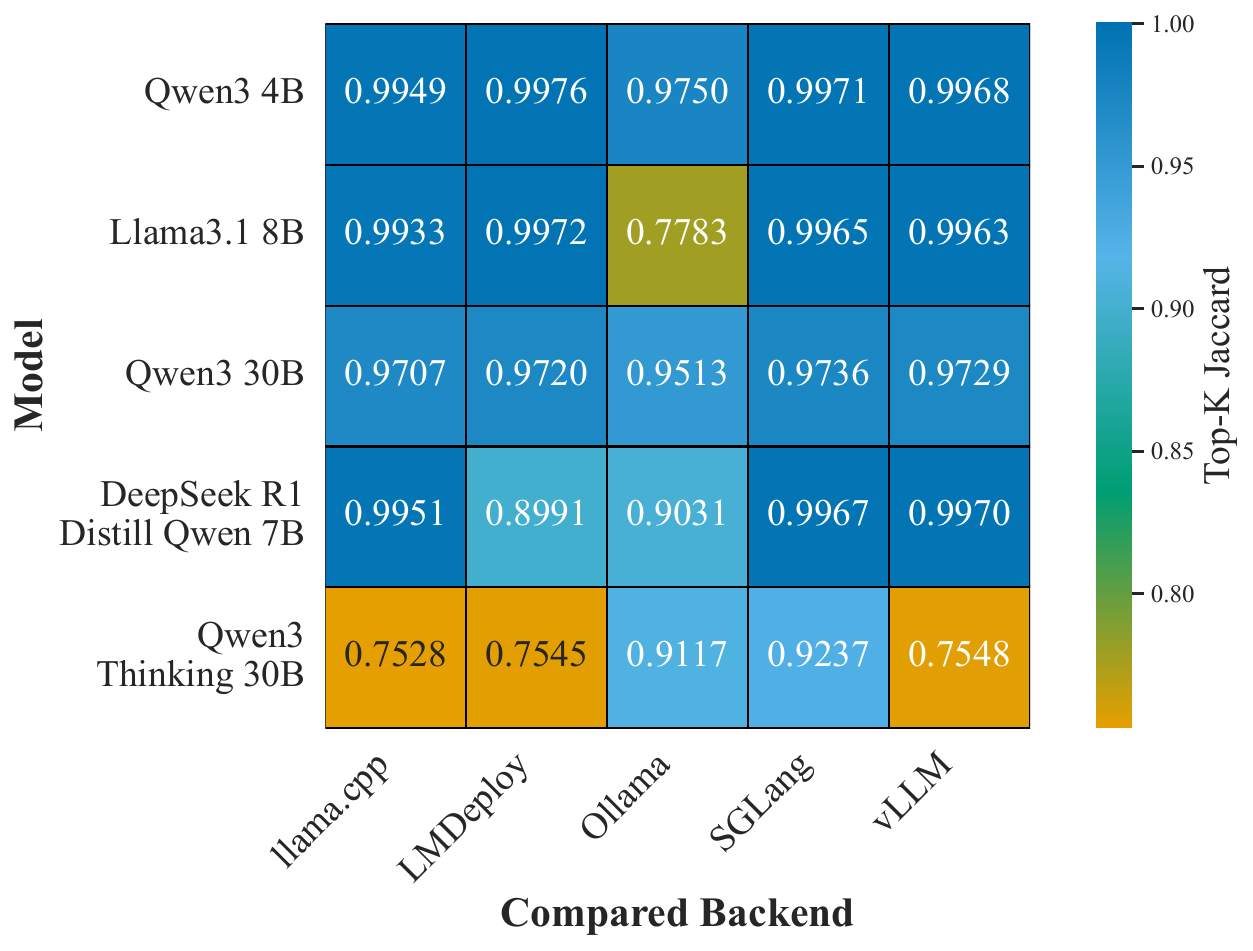}
        \caption{LiveCodeBench}
        \label{fig:jaccard_livecodebench}
    \end{subfigure}
    
    \caption{\textbf{Distribution Stability (Top-5 Jaccard Similarity).} The overlap of the top-5 most probable tokens between the backend and the transformers reference. We expect a similarity score of 1.0, meaning the set of the top-5 token candidates is perfectly identical across both implementations. Lower values indicate that the backend's numerical deviations fundamentally alter the model's candidate pool, bringing entirely different tokens into the top-5 predictions.}
    \label{fig:top_5_jaccard}
\end{figure}

To determine if the backend variance shifts the entire probability distribution or just the absolute top prediction, we calculate the Top-5 Token Jaccard Similarity (Figure~\ref{fig:top_5_jaccard}). This metric measures the overlap of the top-5 candidate tokens between the backend and the reference implementation. While most standard models maintain high similarity, sharp drops in this metric (particularly in reasoning models) indicate that numerical instability is occasionally severe enough to completely reshape the distribution, pushing entirely new tokens in the top-5 candidate pool.

\section{Ablation Studies}\label{app:ablations}
To validate the robustness of our findings and address the real-world implications of backend-induced variance, we conducted four targeted ablation studies. Unless otherwise specified, all ablations evaluate the Llama 3.1 8B and DeepSeek R1 Distill Qwen 7B models across 12 seeds.

\subsection{Safety and Jailbreak Vulnerability}
\label{app:safety}
To assess whether backend-induced variance impacts model alignment, we evaluated vulnerability to adversarial prompts using JailbreakBench. The metric reported is Attack Success Rate (ASR \%), where lower is better. As shown in Table~\ref{tab:ablation_safety}, while Llama 3.1 8B is consistently robust, DeepSeek R1 Distill Qwen 7B's vulnerability fluctuates by 8.9\% depending on the inference engine. This demonstrates that identical model weights can exhibit noticeably distinct safety profiles when deployed with different inference engines. 

\begin{table}[ht]
\centering
\caption{\textbf{Ablation: Safety Vulnerability.} Attack Success Rate (\%) on JailbreakBench (Lower is more robust).}
\label{tab:ablation_safety}
\resizebox{\textwidth}{!}{
\begin{tabular}{l|c|ccccc|c}
\toprule
\textbf{Model} & \textbf{transformers} & \textbf{llama.cpp} & \textbf{LMDeploy} & \textbf{Ollama} & \textbf{SGLang} & \textbf{vLLM} & \textbf{Max - Min} \\
\midrule
Llama 3.1 8B & 02.00 $\pm$ 00.00 & 02.00 $\pm$ 00.00 & 02.00 $\pm$ 00.00 & 02.00 $\pm$ 00.00 & 02.00 $\pm$ 00.00 & 02.00 $\pm$ 00.00 & 00.00 \\
DeepSeek R1 Distill Qwen 7B & 48.90 $\pm$ 00.80 & 48.20 $\pm$ 00.70 & 47.30 $\pm$ 00.70 & \textbf{42.90} $\pm$ 00.30 & 47.10 $\pm$ 01.30 & \underline{51.80} $\pm$ 01.10 & 08.90 \\
\bottomrule
\end{tabular}
}
\end{table}

\subsection{Batching}\label{app:batching}
Inference engines like vLLM and SGLang are specifically designed for high-throughput, concurrent serving environments, relying heavily on mechanisms like continuous batching. To ensure our findings reflect these real-world usage patterns, we ran an ablation using a batch size of 4 on the GSM8K benchmark. We omit llama.cpp and Ollama, because they do not natively support batched generation. As shown in Table~\ref{tab:ablation_batch}, backend variance persists under batched generation. Furthermore, comparing these results to Table~\ref{tab:comparison_experiment} reveals slight numerical shifts within vLLM and SGLang themselves, confirming that batching actively influences the final generation.

\begin{table}[ht]
\centering
\caption{\textbf{Ablation: Batched Generation (Batch Size = 4).} Accuracy (\%) on GSM8K.}
\label{tab:ablation_batch}
\resizebox{\textwidth}{!}{
\begin{tabular}{l|c|ccc|c}
\toprule
\textbf{Model} & \textbf{transformers}  & \textbf{LMDeploy}  & \textbf{SGLang} & \textbf{vLLM} & \textbf{Max - Min} \\
\midrule
Llama 3.1 8B & \underline{84.00} $\pm$ 0.00 & 84.15 $\pm$ 0.00 & \textbf{84.22} $\pm$ 0.02 & \underline{84.00} $\pm$ 0.09 & 0.22 \\
DeepSeek R1 Distill Qwen 7B & \textbf{78.62} $\pm$ 0.00 & \underline{74.07} $\pm$ 0.00 & \textbf{78.62} $\pm$ 0.00 & 78.38 $\pm$ 0.20 & 4.55 \\
\bottomrule
\end{tabular}
}
\end{table}

\subsection{Hardware Variation}\label{app:hardware}
To verify that our observations are not an artifact of our specific NVIDIA H100 (Hopper) setup, we repeated the evaluations on NVIDIA L40 GPUs (Ada Lovelace architecture). Table~\ref{tab:ablation_hardware} shows that variance not only persists but slightly increases on the L40 GPUs. While absolute benchmark scores naturally shift when changing GPU architectures, the relative performance differences induced by the backends remain structurally consistent.

\begin{table}[h!]
\centering
\caption{\textbf{Ablation: Hardware Variation (NVIDIA L40).} Accuracy (\%) on GSM8K.}
\label{tab:ablation_hardware}
\resizebox{\textwidth}{!}{
\begin{tabular}{l|c|ccccc|c}
\toprule
\textbf{Model} & \textbf{transformers} & \textbf{llama.cpp} & \textbf{LMDeploy} & \textbf{Ollama} & \textbf{SGLang} & \textbf{vLLM} & \textbf{Max - Min} \\
\midrule
Llama 3.1 8B & 83.93 $\pm$ 00.00 & 84.00 $\pm$ 00.00 & \textbf{84.76} $\pm$ 00.00 & \underline{73.31} $\pm$ 00.00 & 83.62 $\pm$ 00.00 & 84.08 $\pm$ 00.00 & 11.45 \\
DeepSeek R1 Distill Qwen 7B & 78.01 $\pm$ 00.00 & 78.24 $\pm$ 00.00 & 73.31 $\pm$ 00.00 & \underline{61.71} $\pm$ 00.00 & \textbf{78.92} $\pm$ 00.00 & 78.54 $\pm$ 00.00 & 17.21 \\
\bottomrule
\end{tabular}
}
\end{table}

\subsection{Stochastic Sampling}\label{app:sampling}
While we utilized greedy decoding ($T=0$) to strictly isolate backend differences, real-world deployments often rely on temperature sampling. To confirm that the divergence established at $T=0$ does not disappear when stochastic sampling is introduced, we evaluated GSM8K with temperature $T=0.7$. As seen in Table~\ref{tab:ablation_sampling}, while absolute accuracies drop slightly and standard deviations naturally increase due to sampling randomness, the Max-Min differences remain nearly identical to the greedy decoding setup. Because engines alter token probabilities at the logit level, the underlying distributions from which samples are drawn remain fundamentally different, proving this variance is not an artifact of greedy decoding.

\begin{table}[!h]
\centering
\caption{\textbf{Ablation: Stochastic Sampling (Temperature = 0.7).} Accuracy (\%) on GSM8K.}
\label{tab:ablation_sampling}
\resizebox{\textwidth}{!}{
\begin{tabular}{l|c|ccccc|c}
\toprule
\textbf{Model} & \textbf{transformers} & \textbf{llama.cpp} & \textbf{LMDeploy} & \textbf{Ollama} & \textbf{SGLang} & \textbf{vLLM} & \textbf{Max - Min} \\
\midrule
Llama 3.1 8B & \textbf{82.00} $\pm$ 00.70 & 81.00 $\pm$ 00.70 & 80.90 $\pm$ 01.00 & \underline{73.30} $\pm$ 01.00 & 81.20 $\pm$ 00.80 & 81.10 $\pm$ 00.80 & 08.70 \\
DeepSeek R1 Distill Qwen 7B & \textbf{78.10} $\pm$ 01.00 & 77.40 $\pm$ 01.70 & 70.40 $\pm$ 02.90 & \underline{61.70} $\pm$ 01.10 & 77.90 $\pm$ 00.50 & 76.50 $\pm$ 03.90 & 16.30 \\
\bottomrule
\end{tabular}
}
\end{table}

\section{Root Cause Analysis}\label{app:root_cause}
As established in Section~\ref{sec:root_cause}, several of the large differences in benchmark performance across inference engines can be traced to specific system-level design choices. To isolate and quantify these sources of variance, we conduct targeted ablations on GSM8K and GPQA using Llama 3.1 8B, DeepSeek R1 Distill Qwen 7B, and Qwen3 4B. The complete results are reported in Tables~\ref{tab:root_cause_gsm8k} and~\ref{tab:root_cause_gpqa}.

Because inference backends differ in their architectures, default configurations, and optimization techniques, not every ablation applies to every engine or model. For example, Ollama applied hidden preprocessing and logit-processing defaults, whereas vLLM and SGLang primarily differ through high-throughput execution optimizations. Similarly, hidden BOS insertion is not applicable to Qwen3 4B because its chat template does not use a BOS token.

We broadly categorize these root causes into two groups:
\begin{enumerate}
    \item \textbf{Systematic Engine Defaults:} These are correctable, engine-specific behaviors that alter the prompt or decoding procedure. Hidden BOS insertion and default repetition penalties produce the largest effects in our ablations when applicable. On GSM8K, removing hidden BOS tokens improves accuracy by up to 8.34 percentage points, while disabling Ollama's repetition penalty improves DeepSeek R1 Distill Qwen 7B by 11.67 points. The GPQA experiments show the same mechanisms can remain consequential across tasks, although their magnitude depends on the model and backend.
    \item \textbf{Optimization-Induced Numerical Variance:} Even after prompt construction and decoding defaults are aligned, backend-specific execution paths introduce numerical differences. Prefix caching, CUDA graphs, and custom greedy decoding kernels can alter floating-point computation and subsequent token selection. In contrast to systematic defaults, their effects vary in both magnitude and direction across configurations. In our experiments, these ablations range from no net accuracy change to shifts of 3.54 percentage points on GPQA.
\end{enumerate}

Tables~\ref{tab:root_cause_gsm8k} and~\ref{tab:root_cause_gpqa} report the baseline accuracy and the results after ablating each feature. Importantly, an unchanged aggregate accuracy does not imply identical model behavior: even configurations with $\Delta=0$ exhibit sample-level disagreement with their corresponding baseline.

\begin{table}[!h]
\centering
\caption{\textbf{Root Cause Ablation Results on GSM8K.} Performance impact of isolating specific system-level defaults and hardware optimizations. Metrics are reported as Accuracy (\%). The $\Delta$ columns report the percentage-point change relative to the original configuration.}
\label{tab:root_cause_gsm8k}
\resizebox{\textwidth}{!}{
\begin{tabular}{llccc ccc ccc}
        \toprule
        \multirow{2}{*}{\textbf{Inference Engine}}
        & \multirow{2}{*}{\textbf{Ablated Feature / Optimization}}
        & \multicolumn{3}{c}{\textbf{Llama 3.1 8B}}
        & \multicolumn{3}{c}{\textbf{DeepSeek R1 Distill Qwen 7B}}
        & \multicolumn{3}{c}{\textbf{Qwen3 4B}} \\
        \cmidrule(lr){3-5}
        \cmidrule(lr){6-8}
        \cmidrule(lr){9-11}
        & & Original & Ablated & $\Delta$
          & Original & Ablated & $\Delta$
          & Original & Ablated & $\Delta$ \\
        \midrule

        llama.cpp
        & Disable Prefix Caching
        & 84.15 & 84.08 & $-0.07$
        & 78.24 & 78.70 & $+0.46$
        & 90.45 & 90.67 & $+0.22$ \\

        \midrule
        \multirow{2}{*}{LMDeploy}
        & Patch Argmax Kernel Tie-Breaking
        & 84.53 & 84.08 & $-0.45$
        & 73.54 & 73.62 & $+0.06$
        & 90.45 & 90.45 & $0.00^{*}$ \\

        & Remove Hidden BOS Token
        & -- & -- & --
        & 73.54 & 78.24 & $+4.70$
        & -- & -- & -- \\

        \midrule
        \multirow{3}{*}{Ollama}
        & Disable Prefix Caching
        & 74.30 & 73.24 & $-1.06$
        & 61.87 & 61.41 & $-0.47$
        & 90.75 & 90.52 & $-0.23$ \\

        & Remove Hidden BOS Token
        & 74.30 & 82.64 & $+8.34$
        & 61.87 & 69.22 & $+7.35$
        & -- & -- & -- \\

        & Disable Repetition Penalty$^{\dagger}$
        & 74.30 & 75.97 & $+1.67$
        & 61.87 & 73.54 & $+11.67$
        & 90.75 & 90.75 & $0.00^{*}$ \\

        \midrule
        \multirow{2}{*}{SGLang}
        & Disable Prefix Caching
        & 84.23 & 83.85 & $-0.48$
        & 78.24 & 78.54 & $+0.30$
        & 90.45 & 90.30 & $-0.15$ \\

        & Disable CUDA Graphs
        & 84.23 & 84.23 & $0.00^{*}$
        & 78.24 & 78.24 & $0.00^{*}$
        & 90.45 & 90.45 & $0.00^{*}$ \\

        \midrule
        \multirow{2}{*}{vLLM}
        & Disable Prefix Caching
        & 83.85 & 84.31 & $+0.46$
        & 78.32 & 78.32 & $0.00^{*}$
        & 90.47 & 90.67 & $+0.20$ \\

        & Disable CUDA Graphs
        & 83.85 & 84.00 & $+0.15$
        & 78.32 & 78.47 & $+0.15$
        & 90.47 & 90.60 & $+0.13$ \\

        \bottomrule
        \multicolumn{11}{l}{ $^{\dagger}$ Ollama enforces a default repetition penalty of 1.1; disabling it sets the penalty to 1.0.} \\
        \multicolumn{11}{l}{$^{*}$ A zero net accuracy shift does not imply identical predictions; these configurations still exhibit sample-level disagreement.} \\
    \end{tabular}%
    }
\end{table}

\begin{table}[!h]
\centering
\caption{\textbf{Root Cause Ablation Results on GPQA.} Performance impact of isolating specific system-level defaults and hardware optimizations. Metrics are reported as accuracy (\%). The $\Delta$ columns report the percentage-point change relative to the original configuration. GPQA contains 198 examples, such that one net change in correctness corresponds to approximately 0.51 percentage points.}
\label{tab:root_cause_gpqa}
\resizebox{\textwidth}{!}{
\begin{tabular}{llccc ccc ccc}
        \toprule
        \multirow{2}{*}{\textbf{Inference Engine}}
        & \multirow{2}{*}{\textbf{Ablated Feature / Optimization}}
        & \multicolumn{3}{c}{\textbf{Llama 3.1 8B}}
        & \multicolumn{3}{c}{\textbf{DeepSeek R1 Distill Qwen 7B}}
        & \multicolumn{3}{c}{\textbf{Qwen3 4B}} \\
        \cmidrule(lr){3-5}
        \cmidrule(lr){6-8}
        \cmidrule(lr){9-11}
        & & Original & Ablated & $\Delta$
          & Original & Ablated & $\Delta$
          & Original & Ablated & $\Delta$ \\
        \midrule

        llama.cpp
        & Disable Prefix Caching
        & 23.74 & 23.23 & $-0.51$
        & 31.82 & 31.31 & $-0.51$
        & 38.89 & 38.38 & $-0.51$ \\

        \midrule
        \multirow{2}{*}{LMDeploy}
        & Patch Argmax Kernel Tie-Breaking
        & 25.76 & 25.76 & $0.00^{*}$
        & 35.86 & 32.32 & $-3.54$
        & 35.35 & 34.85 & $-0.50$ \\

        & Remove Hidden BOS Token
        & -- & -- & --
        & 35.86 & 35.86 & $0.00^{*}$
        & -- & -- & -- \\

        \midrule
        \multirow{3}{*}{Ollama}
        & Disable Prefix Caching
        & 22.22 & 23.23 & $+1.01$
        & 27.27 & 30.81 & $+3.54$
        & 37.88 & 36.87 & $-1.01$ \\

        & Remove Hidden BOS Token
        & 22.22 & 26.77 & $+4.55$
        & 27.27 & 29.29 & $+2.02$
        & -- & -- & -- \\

        & Disable Repetition Penalty$^{\dagger}$
        & 22.22 & 24.24 & $+2.02$
        & 27.27 & 28.79 & $+1.52$
        & 37.88 & 37.88 & $0.00^{*}$ \\

        \midrule
        \multirow{2}{*}{SGLang}
        & Disable Prefix Caching
        & 23.23 & 23.74 & $+0.51$
        & 31.31 & 30.81 & $-0.50$
        & 34.85 & 35.86 & $+1.01$ \\

        & Disable CUDA Graphs
        & 23.23 & 24.75 & $+1.52$
        & 31.31 & 31.31 & $0.00^{*}$
        & 34.85 & 34.85 & $0.00^{*}$ \\

        \midrule
        \multirow{2}{*}{vLLM}
        & Disable Prefix Caching
        & 25.76 & 25.76 & $0.00^{*}$
        & 33.84 & 34.34 & $+0.50$
        & 34.81 & 34.85 & $+0.04$ \\

        & Disable CUDA Graphs
        & 25.76 & 25.25 & $-0.51$
        & 33.84 & 32.32 & $-1.52$
        & 34.81 & 35.86 & $+1.05$ \\

        \bottomrule
        \multicolumn{11}{l}{ $^{\dagger}$ Ollama enforces a default repetition penalty of 1.1; disabling it sets the penalty to 1.0.} \\
        \multicolumn{11}{l}{$^{*}$ A zero net accuracy shift does not imply identical predictions; these configurations still exhibit sample-level disagreement.} \\
    \end{tabular}%
    }
\end{table}

\pagebreak

\section{Paper Survey Methodology and Artifacts}\label{app:paper_survey_methodology}
To conduct the large-scale literature survey detailed in Section~\ref{sec:paper_survey}, we employed a multi-stage automated extraction pipeline utilizing an LLM-as-a-judge. This section documents the exact methodology and artifacts used to ensure the reproducibility of our survey. First, we outline the logic of our extraction pipeline. Then, we provide the specific Python lists of target keywords and dependency files used for the initial filtering and code repository validation. Finally, we include the exact prompts provided to the LLM judge.

\subsection{Pipeline Methodology}\label{app:paper_survey_pipeline}
\textbf{Corpus Collection.} We retrieved accepted papers from NeurIPS, ICML, and ICLR through the OpenReview API and from ACL and EMNLP through the ACL Anthology API, covering publication years 2023-2025. This produced the initial corpus of more than 35,000 PDFs used by the filtering pipeline below.\\
To efficiently and accurately process this initial corpus, our automated extraction pipeline was executed in sequential stages:
\begin{enumerate}[left=0pt, itemsep=0pt, topsep=0pt]
    \item \textbf{Keyword Pre-Filtering:} We first applied a heuristic pre-filter, since running an LLM judge over the full corpus was computationally expensive. We scanned the raw extracted text (using \texttt{pymupdf} Python library) of all PDFs for specific terms related to open-weight models and local execution (see Section~\ref{app:paper_survey_keywords}). Only papers containing at least one of these keywords advanced to the LLM judge. 
    \item \textbf{LLM Prompting Strategy:} We utilized Qwen3-235B-A22B-Instruct-2507-AWQ (using vLLM and greedy decoding) as our LLM judge, processing the full text of each paper. To maximize accuracy and minimize hallucinations, the judge was prompted using a Chain-of-Thought (CoT) approach. As seen in the prompts, the model was instructed to first output a \texttt{thought\_process} explaining its reasoning based on the provided inclusion/exclusion criteria, before outputting the final classification or extracted text.
    \item \textbf{Structured Output:} To allow for automated parsing of the LLM's decisions, the judge was strictly prompted to return responses in a valid JSON format. This allowed our evaluation scripts to programmatically route papers through the subsequent \emph{Code Extraction} and \emph{Engine Extraction} stages based on the boolean flags generated during the \emph{Relevance Filtering} stage.
\end{enumerate}

\subsection{Filtering Keywords}\label{app:paper_survey_keywords}
\begin{tcolorbox}[width=\textwidth, enhanced jigsaw,breakable,pad at break*=1mm,
  colback=blue!5!white,colframe=blue!75!black, title=Relevance Filtering Keywords, center]
\small
\begin{verbatim}
KEYWORDS = [
    "large language model", "large language models",
    "LLM", "LLMs",
    "language model", "language models",
    
    "Transformer", "Transformers", 
    "Transformer-based",
    
    "generative AI",
    "foundation model", "foundation models",
    "in-context learning",
    "chain-of-thought",
    
    "GPT", "Llama", "Mistral", "Falcon", "Qwen", "DeepSeek", 
]
\end{verbatim}
\end{tcolorbox}

\subsection{LLM-Judge Prompts}\label{app:paper_survey_judge_prompts}
\begin{tcolorbox}[width=\textwidth, enhanced jigsaw,breakable,pad at break*=1mm,
  colback=blue!5!white,colframe=blue!75!black, title=Relevance Filtering Judge Prompt, center]
\tiny
\begin{verbatim}
You are an expert NLP researcher conducting a systematic survey on "LLM Inference Backends."
Your task is to analyze the text of a research paper and determine if it is RELEVANT for a study on how inference 
engines (like vLLM, llama.cpp, etc.) affect the performance of Open-Source/Local **Text-Only** Large Language Models.
### STRICT EXCLUSION CRITERIA (CHECK THESE FIRST)
Mark as "relevant": false if ANY of the following apply, even if other criteria are met:
1.  **Multimodal Inputs (Vision/Audio)**:
   *   **The Paper uses Images, Video, or Audio as input.**
   *   **VLMs are EXCLUDED**, even if they use a Llama/Qwen backbone.
   *   **Diffusion/Generative image models are excluded**
   *   *Excluded Models*: LLaVA, Qwen-VL, GPT-4V, Phi-Vision, CLIP, MiniCPM-V, BakLLaVA, Yi-VL.
   *   *Reasoning*: The inference stack for VLMs involves visual encoders/projectors, which is outside the scope of 
   text-only inference backends.
2.  **Non-Generative Architectures**:
   *   **Topic Models / Clustering**: Papers focusing on extracting topics (LDA, BERTopic, Autoencoders) without
   autoregressive generation.
   *   **Embeddings Only**: Papers that only use the model to generate vector embeddings (hidden states) for 
   retrieval/search, without decoding text.
   *   **Encoder-Only / Autoencoders**: BERT, RoBERTa, DeBERTa, VAEs.
   *   **Non-Transformers**: RNNs, LSTMs, SSMs (Mamba/RWKV), etc. *unless* comparing against Transformer LLMs.
3. **Purely Proprietary/Black-Box**: The paper ONLY uses closed-source models without comparing them to local models.
   *   *Exclusion List*: GPT-3.5, GPT-4, GPT-4o, o1, GPT-5, OpenAI, Claude (Sonnet/Opus/Haiku), Gemini (Pro/Ultra), 
   PaLM, Grok (proprietary versions) etc.
   *   *Exception*: If the paper compares GPT-4 vs. Llama 2, it is RELEVANT.
4.  **Secondary Analysis of Pre-Generated Data (PASSIVE USAGE)**:
   *   **CRITICAL EXCLUSION**: If the authors use an *existing dataset* (e.g., a corpus, a benchmark, or human-eval 
   data) where the text was generated by LLMs in a *previous study*, this paper is **IRRELEVANT**.
   *   *Example of Exclusion*: "We analyze the *EMTeC corpus* (Smith et al.), which contains text generated by 
   Llama-2." (Here, the *current* authors did not run Llama-2; Smith et al. did).
   *   *Reasoning*: We are studying the inference engine used *by these authors*. If they are analyzing downloaded 
   data, they are not running an inference backend.
### INCLUSION CRITERIA (MUST MEET ALL) 
To be marked as "relevant": true, the paper must meet these conditions:
1.  **Task = Autoregressive Text Generation**:
    *   The model must receive **Text** as input and generate **Text/Code** (or logits for text tokens) as output.
    *   The mechanism must be next-token prediction (Transformer Decoder).
2.  **Model = Open-Weights / Local**:
    *   The authors must utilize models where weights are publicly available or can be hosted locally. 
    *   *Examples*: Llama (1, 2, 3), Mistral, Mixtral, Qwen (Text-only), DeepSeek (Text-only), Gemma, Phi, Yi, 
    Falcon, OPT, Dolphin, Kimi, Vicuna, Alpaca, Pythia, BLOOM, OLMo, Solar, StarCoder.
3.  **Action = Running Inference**:
   *   The authors must **actively execute** the model themselves during the course of the study.
   *   This includes:
      *   Running the model to generate *new* responses.
      *   Running the model to calculate perplexity/logits on a dataset.
      *   Running the model to benchmark speed/latency.
   *   *Note*: Papers that Fine-Tune (SFT/RLHF/GRPO etc.) are RELEVANT if they subsequently evaluate the model 
   using inference (calculating accuracy, perplexity, or generating responses).
   *   *Note*: Usage via APIs (e.g., Together AI, Anyscale) is RELEVANT if the underlying model is open-weights 
   (e.g., calling Llama-3-70B via API).
   *   *Note:* Papers focused on "LLM-as-a-judge" or "Synthetic Data Generation" using open models ARE relevant.
### OPERATIONAL GUIDELINES
1. **Robustness to Artifacts**: The input text is extracted from PDFs and may contain OCR errors, headers/footers, 
broken lines, or merged words (e.g., "Lla ma-2", "Hugging Face", "Q wen"). Look past these structural issues to 
understand the semantic content.
2. **Model Family Inheritance**: Use the model's name to infer its nature. 
   - If a model is unknown to you (e.g., "Llama-4" or "Mistral-Next") but shares a name with a known open-source 
   family (Llama, Mistral, Qwen, etc.), **assume it is open-source**.
   - Conversely, if it shares a name with a proprietary family (e.g., "GPT-5", "Claude-Next"), assume it is excluded.
3. **Inference Engine Agnosticism**: 
   - **Do not look for specific engine names** (like vLLM, llama.cpp, SGLang) to determine relevance. 
   - Many authors fail to report their backend. If the paper *uses* a relevant model (e.g., Llama 2) for inference, 
   it is **RELEVANT**, regardless of whether they mention the software stack used to run it. 
4. **Non-Exclusive Examples**: The inclusion/exclusion model lists provided above are **representative samples**, not 
exhaustive lists. If a paper uses a model not listed (e.g., "MiniCPM" or "XVerse"), use your judgment: if it is an 
open-weights generative transformer, include it.
5. **Knowledge Cutoff & New Models**: You may encounter models released after your training data cutoff. **Do not 
hallucinate**. Instead, look for context clues in the text to classify them.
   - *Clues for Relevance*: "weights released," "available on GitHub/HuggingFace," "reproduced locally," "7B 
   parameters."
   - *Clues for Exclusion*: "proprietary model," "image generation," "diffusion process."
6. **Indirect Citations (Reference Lookup)**: If the authors refer to a model only by citation (e.g., "We utilize 
the model proposed by Touvron et al. [15]" or "the model from [1]"), you **MUST** look at the 
References/Bibliography section at the end of the text to identify the model. If citation [15] is the "Llama 2" 
paper, then the paper is RELEVANT.
7. **Burden of Proof (Uncertainty = Reject)**: You must find **positive evidence** of the criteria above. If the 
text is too vague, lacks sufficient information, or you are unsure, mark it as **"relevant": false**.
8.  **Dataset Origin vs. Experimentation (The "Created By" Check)**: 
    - Pay close attention to grammar. If the text says: *"We use Data X (Author, Year), which was created using 
    Model Y"*, the paper is **NOT RELEVANT** (unless they *also* run Model Y separately).
    - If the text says: *"We used Model Y to create Data X"*, the paper is **RELEVANT**.
---
### INPUT TEXT
<paper_text>
{full_pdf_text_extracted}
</paper_text>
---
### FINAL INSTRUCTION
Based on the text above, determine if this paper is RELEVANT.
Respond with valid JSON only:
{
  "thought_process": "Brief explanation. Did you find specific open model names? Did they run inference?",
  "relevant": boolean
}
\end{verbatim}
\end{tcolorbox}

\begin{tcolorbox}[width=\textwidth, enhanced jigsaw,breakable,pad at break*=1mm,
  colback=blue!5!white,colframe=blue!75!black,title=Engine Extraction Judge Prompt, center]
\tiny
\begin{verbatim}
You are an expert Systems and Reproducibility Researcher analyzing a machine learning paper.
Your goal is to extract the **Inference Engine(s)** or **Backend(s)** used to actively execute **Open-Weight / Local 
Text-Only Models**.
### 1. TARGET SCOPE (READ CAREFULLY)
You must filter your extraction based on the **Model** and **Task**.
*   **INCLUDE (Target Models)**: Open-Weights or Local models (e.g., Llama, Mistral, Qwen, DeepSeek, Gemma, Phi, 
Falcon, Yi, OLMo).
    *   *Unknown Models*: If a model name is unfamiliar (e.g., "X-7B"), look for context clues ("weights on 
    HuggingFace", "locally hosted"). If it looks like an open generative transformer, count it.
*   **INCLUDE (Target Task)**: Autoregressive Text Generation / Code Generation.
*   **EXCLUDE (Do not extract backends for these)**:
    *   **Proprietary Models**: If the paper runs GPT-4, Claude, or Gemini, **IGNORE** the API/Backend used for them. 
    We only care about the open-source side.
    *   **Multimodal (VLMs)**: LLaVA, Qwen-VL, CLIP. (Inference stacks for vision differ from text-only stacks).
    *   **Non-Generative**: BERT, Encoders, Embeddings-only.
**Example of Mixed Usage:**
If a paper says: *"We compared GPT-4 (via OpenAI API) against Llama-3 (running on vLLM)."*
-> **Result**: Extract `vLLM`. Ignore `OpenAI API`.
### 2. DEFINITION: WHAT IS AN INFERENCE ENGINE?
An inference engine is the specific software stack that manages the model's weights and executes the forward pass 
(generation). It is distinct from the model itself (e.g., "Llama-3" is a model; "vLLM" is the engine).
We categorize engines into three types:
1.  **Self-Hosted Libraries**: Software running on the user's hardware (e.g., `vLLM`, `llama.cpp`, `SGLang`, 
`HuggingFace Transformers`, `TGI`, `LMDeploy`, `TensorRT-LLM`).
2.  **Managed Inference Platforms**: APIs serving open-weight models (e.g., `Together AI`, `Fireworks AI`, 
`RunPod Serverless`).
3.  **Aggregators**: Routers that sit in front of providers (e.g., `OpenRouter`, `LiteLLM`).
### 3. KNOWN ENGINE LIST (Reference Only)
Use this list to help identify potential candidates, but **do not limit yourself to it**. Context matters more than 
the list.
<known_engines>
{known_engines_list}
</known_engines>
### 4. CRITICAL LOGIC: ACTIVE EXECUTION vs. PASSIVE CITATION
Just because an engine is mentioned does not mean it was used.
*   **TRUE (Used)**:
    *   "We generated responses using **vLLM**."
    *   "Latency was measured on **llama.cpp**."
    *   "Models were deployed using **TGI**."
    *   "We use the standard **HuggingFace** implementation."
*   **FALSE (Reference/Comparison)**:
    *   "vLLM [15] is a popular system." (Background info).
    *   "We compare our method against the numbers reported by SGLang." (They didn't run SGLang; they just cited 
    numbers).
    *   "We used the dataset from Smith et al., which was generated using vLLM." (Passive usage).
### 5. PRECISE NAMING & UNKNOWN LIBRARIES
**This is the most critical step for new or specialized tools.**
1.  **Do Not Over-Normalize**: Many libraries have similar names. Do not merge them unless they are aliases.
    *   *Example:* If the text says `FastTransformer`, do NOT map it to `transformers`. Report `FastTransformer`.
    *   *Rule:* Only map generic terms like "HuggingFace", "HF", or "AutoModel" to `transformers`. If a specific, 
    distinct library name is used (even if it contains the word "Transformer"), **extract the exact name**.
2.  **Unknown/New Libraries**: The authors may use a library not in your known list or one released after your 
knowledge cutoff.
    *   *Rule:* If the text explicitly states a software tool was used for inference/execution, **extract it**, even 
    if you have never heard of it. Trust the text.
### 6. ROBUSTNESS & NORMALIZATION
*   **OCR Artifacts**: Fix broken text. `v LLM` -> `vLLM`, `llama . cpp` -> `llama.cpp`.
*   **Ambiguity**:
    *   "PyTorch" / "Native PyTorch": If they wrote a custom loader/engine, mark as `Custom/PyTorch`.
    *   "JAX": If they wrote a custom loader/engine, mark as `Custom/JAX`.
    *   "TensorFlow": If they wrote a custom loader/engine, mark as `Custom/TensorFlow`.
## SEARCH STRATEGY (Where to look)
The engine name is most likely found in one of these locations:
*   **Experimental Setup / Implementation Details**: The most likely location.
*   **Footnotes**: Authors often bury the backend version or name here (e.g., "We used vLLM v0.2.3").
*   **Appendix**: Look for "Compute Resources" or "Hyperparameters".
*   **Code Snippets**: Look for imports like `from vllm import LLM` or `import sglang`.
However, do not limit yourself to search for the engine only in those sections. Use the full paper.
### OUTPUT FORMAT
Respond with valid JSON only.
{
  "thought_process": "Step 1: Identify open-weight models used (e.g., 'They used Llama-2'). Step 2: Look for the 
  execution software for THOSE models. Step 3: Verify active execution (not just citation). Step 4: Check if this 
  software is a distinct library. Quote the relevant sentence.",
  "backend_reported": boolean, // true if they explicitly name the software stack used for open models
  "backends_found": [string, string] // List of clean names (e.g., ["vLLM", "transformers"]). Empty [] if none found.
}
### INPUT TEXT
<paper_text>
{full_pdf_text_extracted}
</paper_text>
\end{verbatim}
\end{tcolorbox}

\begin{tcolorbox}[width=\textwidth, enhanced jigsaw,breakable,pad at break*=1mm,
  colback=blue!5!white,colframe=blue!75!black,title=Code Extraction Judge Prompt, center]
\tiny
\begin{verbatim}
You are an expert Reproducibility Reviewer for a top-tier Machine Learning conference.

Your sole objective is to analyze the full text of a research paper and locate the **official code repository** 
provided by the authors.

### THE CHALLENGE
You are working with raw text extracted from a PDF. This text often contains errors, such as:
*   **Broken URLs**: `git hub . com / user / repo` or `https://github.com/ \n user/repo`.
*   **Merged Text**: `code is available atgithub.com/user/repo`.
*   **Hidden Links**: Links might be in footnotes, references, or the abstract.

**You must robustly scan the text, identify the link, clean it, and verify it.**

### SEARCH STRATEGY (Where to look)
The code link is almost always found in one of these locations. Check them mentally in this order:
1.  **Abstract**: specifically the very last sentence.
2.  **Introduction**: specifically in the "Contributions" list or the final paragraph.
3.  **Footnotes**: Look for text like "See footnote 1" or "[1]" near the mention of code.
4.  **Methodology header**: Sometimes listed as "Implementation Details".
5.  **Conclusion**: A section named "Reproducibility" or "Data Availability".
6.  **References/Bibliography**: Rarely, authors cite their own code as a bibliography entry (e.g., "Source Code 
[25]").

### CRITICAL DECISION LOGIC

**1. Verification of Ownership (The "Author" Check)**
You must distinguish between **Own Work** and **Prior Work**.
*   **RELEVANT (True)**: "We release our code at...", "The official implementation is available at...", "Project 
page: [URL]", "Code: [URL]", "Our code is open-sourced."
*   **IRRELEVANT (False)**: "We used the implementation from [Citation]", "Built upon the codebase of [Citation]", 
"We use the HuggingFace library", "Model weights are available at [URL]" (if strictly weights only).

**2. Handling "Coming Soon" (The Promise Check)**
*   **False**: "Code will be released upon acceptance", "We plan to release code soon.", "Code available upon 
request."
*   **True**: Anonymized repositories used for review (e.g., `anonymous.4open.science`) ARE valid code reporting.
*   **True (Specific Link Provided)**: If the text provides a **SPECIFIC URL**, mark this as `true` **even if the 
text uses future tense** (e.g., "Code *will be* released at...", "We *plan to* release code at...").

**3. Robustness to OCR/Parsing Artifacts (Reconstruction)**
*   PDF parsing often breaks URLs with spaces, hyphens, or newlines.
*   *Task:* If you find a broken URL in the text that points to the author's code, **you must clean it** (remove 
spaces, fix formatting) and return the corrected URL in the JSON output.

**4. Valid Targets**
*   **Repositories**: GitHub, GitLab, Bitbucket, etc.
*   **Project Pages**: (e.g., `github.io`, `site.net`) ARE valid if the text implies code is linked there.
*   **Anonymous Links**: `anonymous.4open.science`, Dropbox, Google Drive (if explicitly stated as the code release).
*   **HuggingFace**: If the authors link to a HuggingFace repository that clearly contains the *code implementation* 
and not just model weights/checkpoints, accept it. If unsure, prioritize GitHub.

### URL RECONSTRUCTION INSTRUCTIONS
The text extraction may insert spaces or newlines into URLs.
*   *Raw Text:* "g it hub . c om / my lab / my repo"
*   *Your Output:* "https://github.com/mylab/myrepo"
*   **Action**: You must intelligently remove whitespace and fix formatting to produce a valid URL string.

### OUTPUT FORMAT
Respond with valid JSON only.

{
  "thought_process": "Analyze the text. 1) Did they mention releasing code? 2) Is the link for THEIR code or a 
  library? 3) Does the link look like a repository? Briefly explain your reasoning regarding ownership and text 
  location.",
  "code_reported": boolean, // true if they linked their own code
  "official_url": string OR null // The best specific URL found. If null, return null. CLEAN THE URL (remove spaces
  /newlines) before returning.
}

---
### INPUT DATA
<paper_text>
{full_pdf_text_extracted}
</paper_text>
\end{verbatim}
\end{tcolorbox}

\subsection{Dependency File Patterns}\label{app:dep_files}

\begin{tcolorbox}[width=\textwidth, enhanced jigsaw,breakable,pad at break*=1mm,
  colback=blue!5!white,colframe=blue!75!black,title=Dependency Files, center]
\small
\begin{verbatim}
DEPENDENCY_FILES = [
    "requirements.txt", "requirements.pip", "pyproject.toml", "poetry.lock", 
    "pipfile", "pipfile.lock", "setup.py", "setup.cfg", "tox.ini",
    
    "environment.yml", "environment.yaml", "meta.yaml",
    
    "dockerfile", "docker-compose.yml", "docker-compose.yaml", "containerfile", 
    "devcontainer.json",
    
    "cargo.toml", "cargo.lock", "cmakelists.txt", "makefile", "package.json", 
    "yarn.lock", "pnpm-lock.yaml", "project.toml", "manifest.toml",
    
    "install.sh", "setup.sh", "build.sh"
]
\end{verbatim}
\end{tcolorbox}

\section{Manual Verification of Judge Results}\label{app:verification}
To assess the reliability of our automated pipeline, we conducted a manual verification on a random subset of the corpus. We selected 50 papers for each of the three processing stages (Relevance Filtering, Code Extraction, and Engine Extraction) resulting in a total of 150 manually audited papers. We compared our manual classification against the LLM judge's output to calculate agreement rates and analyze failure modes:
\begin{itemize}[left=0pt, itemsep=0pt, topsep=0pt]
    \item \textbf{Relevance Filtering (88\,\% Agreement):} The primary source of disagreement was papers utilizing unknown or unpopular open-weight models that the judge failed to recognize. Additionally, the judge occasionally missed relevant papers where LLM usage was mentioned exclusively in a specific subsection of the appendix rather than the main body.
    \item \textbf{Code Extraction (96\,\% Agreement):} The few discrepancies arose from two specific scenarios: cases where the judge interpreted a textual ``promise to share code in the future'' as an existing repository, and PDF parsing issues where valid repository links were embedded in a format our parser could not extract.
    \item \textbf{Engine Extraction (94\,\% Agreement):} Disagreements in this stage were primarily due to false positives where the judge flagged low-level kernel libraries as full inference engines. One error also stemmed from the usage of niche libraries not known to the judge.
\end{itemize}

\section{Impact Statement}\label{app:impact_statement}
This paper aims to improve the scientific quality and reproducibility of LLM evaluations. By quantifying the numerical instability introduced by different inference backends, we highlight a critical blind spot in current benchmarking practices. The primary positive impact of this work is to encourage more transparent reporting standards, ensuring that claims of "State-of-the-Art" performance are statistically significant rather than artifacts of system optimizations.

On a broader societal level, this work has implications for AI safety and reliability. As we demonstrate, optimization techniques can alter model behavior. A model aligned for safety in a development environment may exhibit divergent, potentially unsafe behaviors when deployed on high-throughput inference engines. Identifying and mitigating this source of variance is essential for the safe integration of LLMs into critical domains such as healthcare and finance.

\section{Model and Dataset Licenses}\label{app:licenses}
Table~\ref{tab:licenses} summarizes the licenses and terms of use associated with the models and datasets used in our study.

\begin{table*}[h!]
\centering
\caption{LLMs and Datasets used in this paper alongside their licenses.}\label{tab:licenses}
\begin{tabular}{ll}
\toprule
\textbf{Model/Dataset}&\textbf{License} \\
\midrule
meta-llama/Meta-Llama-3.1-8B-Instruct~\cite{grattafiori2024llama3herdmodels} &  Llama Community License \\
Qwen/Qwen3-4B-Instruct-2507~\cite{qwen3} &  Apache-2.0 \\
Qwen3-30B-A3B-Instruct-2507~\cite{qwen3} & Apache-2.0 \\
deepseek-ai/DeepSeek-R1-Distill-Qwen-7B~\cite{Guo_2025} & MIT License \\
Qwen/Qwen3-30B-A3B-Thinking-2507~\cite{qwen3} & Apache-2.0 \\
AIDXteam/Qwen3-235B-A22B-Instruct-2507-AWQ~\cite{Qwen3-235B} & Not specified (base model: Apache-2.0) \\
gpt-4o-mini-2024-07-18~\cite{gpt4omini}\footnotemark & OpenAI Terms of Use \\
GSM8K~\cite{Cobbe2021TrainingVT} & MIT License \\
GPQA Diamond~\cite{rein2024gpqa} & CC BY 4.0 \\
SimpleQA Verified~\cite{haas2025simpleqaverifiedreliablefactuality} & Apache-2.0 \\
LiveCodeBench v6~\cite{jain2024livecodebench} & CC (exact variant not specified) \\
JailbreakBench~\cite{chao2024jailbreakbench} & MIT License \\
\bottomrule
\end{tabular}
\end{table*}

\footnotetext[3]{Used as LLM judge for SimpleQA and JailbreakBench}